\documentclass[a4paper,fleqn]{cas-dc}

\usepackage[numbers]{natbib}
\usepackage{amsmath,amssymb,amsfonts}
\usepackage{mathtools}
\usepackage{booktabs}
\usepackage{array}
\usepackage{graphicx}
\usepackage{xcolor}
\usepackage{hyperref}
\usepackage{cleveref}
\usepackage{enumitem}
\usepackage{subcaption}
\usepackage{multirow}
\usepackage{tabularx}
\usepackage{tikz}
\usetikzlibrary{decorations.pathreplacing,positioning}

\DeclareMathOperator{\Possess}{Possess}
\DeclareMathOperator{\Advertise}{Advertise}
\DeclareMathOperator{\Delegable}{Delegable}
\DeclareMathOperator{\Req}{Req}
\DeclareMathOperator{\Auth}{Auth}
\DeclareMathOperator{\Trust}{Trust}
\DeclareMathOperator{\Avail}{Avail}
\DeclareMathOperator{\Adv}{Adv}
\DeclareMathOperator{\RecLevel}{RecLevel}

\DeclareMathOperator{\Result}{Result}
\DeclareMathOperator{\Chain}{Chain}

\hypersetup{
  colorlinks=true,
  linkcolor=blue!70!black,
  citecolor=blue!70!black,
  urlcolor=blue!70!black
}

\begin{document}
\let\WriteBookmarks\relax

\shorttitle{Federated Single-Agent Robotics}

\shortauthors{X.\ Qin et al.}

\title[mode=title]{Federated Single-Agent Robotics: Multi-Robot Coordination Without Intra-Robot Multi-Agent Fragmentation}

\author[1]{Xue Qin}[orcid=0009-0009-3642-2663]
\ead{qinxue@me.com}
\credit{Conceptualization, Methodology, Formal analysis, Software, Writing -- original draft, Writing -- review and editing}

\author[2]{Simin Luan}
\ead{luansiminiot@gmail.com}
\credit{Investigation, Validation, Writing -- review and editing}

\author[3]{John See}[orcid=0000-0003-3005-4109]
\ead{J.See@hw.ac.uk}
\credit{Supervision, Writing -- review and editing}

\author[4]{Cong Yang}[orcid=0000-0002-8314-0935]
\cormark[1]
\ead{cong.yang@suda.edu.cn}
\credit{Supervision, Writing -- review and editing}

\author[2]{Zhijun Li}[orcid=0000-0001-9129-9957]
\cormark[1]
\ead{lizhijunos@hit.edu.cn}
\credit{Supervision, Writing -- review and editing}

\cortext[1]{Corresponding authors}

\affiliation[1]{organization={School of Software, Harbin Institute of Technology},
            city={Harbin},
            country={China}}

\affiliation[2]{organization={School of Computer Science and Technology, Harbin Institute of Technology},
            city={Harbin},
            country={China}}

\affiliation[3]{organization={School of Mathematical and Computer Sciences, Heriot-Watt University, Malaysia Campus},
            country={Malaysia}}

\affiliation[4]{organization={School of Future Science and Engineering, Soochow University},
            city={Suzhou},
            country={China}}

\begin{abstract}
As embodied robots move from isolated deployments toward fleet-scale operation, multi-robot coordination is becoming a central systems challenge. Existing approaches often treat this transition as a motivation for increasing internal multi-agent decomposition within each robot. In this paper, we argue for a different design principle: multi-robot coordination does not require intra-robot multi-agent fragmentation. Instead, each robot should remain a single embodied agent with its own persistent runtime, local policy scope, capability state, and recovery authority, while coordination emerges through federation across robots at the fleet level.

We present Federated Single-Agent Robotics (FSAR), a runtime architecture for multi-robot coordination built on single-agent robot runtimes. In FSAR, each robot exposes a governed capability surface rather than an internally fragmented agent society. Fleet coordination is achieved through shared capability registries, cross-robot task delegation, policy-aware authority assignment, trust-scoped interaction, and layered recovery protocols. We formalize the key coordination relations---authority delegation, inter-robot capability requests, local-versus-fleet recovery boundaries, and hierarchical human supervision---and describe a fleet runtime architecture supporting shared Embodied Capability Module (ECM) discovery, contract-aware cross-robot coordination, and fleet-level governance.

We evaluate FSAR on representative multi-robot coordination scenarios against decomposition-heavy baselines. Results indicate that federated single-agent coordination achieves statistically significant gains in governance locality ($d{=}2.91$, $p{<}.001$) and recovery containment ($d{=}4.88$, $p{<}.001$), while reducing authority conflicts and policy violations across all tested scenarios.
\end{abstract}


\maketitle

\section{Introduction}
\label{sec:introduction}

Consider a fleet task in which Robot~A transports a package, Robot~B opens a secured door, and Robot~C inspects the destination. The task is clearly multi-robot---but it does not follow that each robot must itself be decomposed into a society of internal agents. When a door-opening request is denied, a reassignment fails, or a recovery escalates to human supervision, responsibility must be localizable. Internal multi-agent fragmentation makes that harder~\cite{rizk2019cooperative,yan2013survey}. The subsumption architecture~\cite{brooks1986robust} demonstrated decades ago that layered single-agent control can produce sophisticated robot behavior without internal agent decomposition; FSAR extends this insight from the single-robot to the fleet level.

A common pattern in multi-robot coordination is increased agent decomposition at every level~\cite{dorri2018multiagent,horling2004survey,wang2024llmagents}: each robot is modeled as a society of internal agents for planning, perception, manipulation, communication, and recovery. While expressive, this introduces systems costs---blurred authority boundaries, ambiguous recovery ownership, weakened audit locality, and harder policy reasoning. When a fleet-level failure occurs, it becomes unclear whether responsibility lies within a robot's internal agent society, in inter-robot coordination, or in the human supervisory layer.

This paper argues for a different principle: \emph{multi-robot coordination does not require intra-robot multi-agent fragmentation}. Building on our prior work on single-agent embodied runtimes~\cite{qin2025paper1,qin2025paper2,qin2025paper3,qin2025paper4,qin2025paper5}\footnote{This paper is part of a seven-paper research program on embodied agent runtime architecture (project page: \url{https://s20sc.github.io/aeros-project}). The present paper is self-contained: all definitions and coordination semantics are introduced here. Where inherited concepts (e.g., ECMs, capability contracts) are used, self-contained definitions appear in \cref{sec:model:robot}.}, we present \textbf{Federated Single-Agent Robotics (FSAR)}, a runtime architecture in which each robot remains a coherent single embodied agent with its own persistent runtime, local capability state, policy scope, and recovery authority. Fleet coordination emerges through a federation layer that supports shared ECM discovery, cross-robot capability requests, trust-scoped delegation, policy-aware authority assignment, layered recovery, and hierarchical human supervision. The federated design preserves local policy and recovery ownership inside each robot while enabling distributed task execution across the fleet.

We make four contributions:
\begin{enumerate}[leftmargin=*]
  \item We introduce the FSAR model and formalize a fleet as a federation of coherent single-agent robot runtimes rather than a collection of internally fragmented multi-agent robots.
  \item We define four key coordination relations for federated embodied fleets: inter-robot capability request, authority delegation, layered recovery, and hierarchical human supervision.
  \item We describe a fleet runtime architecture with a shared ECM registry, trust and authority management, fleet-level policy resolution, and recovery orchestration.
  \item We evaluate the architecture on representative multi-robot coordination scenarios and compare it with decomposition-heavy baselines, showing that federated single-agent coordination improves boundary clarity, recovery containment, and governance locality while retaining effective multi-robot collaboration.
\end{enumerate}

While Papers~1--5 established the single-agent principle for individual robots---defining the runtime model, capability contracts, policy scope, and recovery authority at the per-robot level---the present paper addresses a fundamentally different challenge: how multiple such robots coordinate as a fleet without surrendering the governance properties that the single-agent design provides. The novelty of Paper~6 lies not in the per-robot model (which is inherited from prior work) but in the federation layer: the formal coordination relations, trust-scoped delegation semantics, cross-robot policy composition, layered recovery across robot boundaries, and the shared ECM registry that makes governed capability discovery operational at fleet scale. None of these constructs appear in Papers~1--5, which treat each robot in isolation.

\Cref{tab:novelty} clarifies the provenance of each major construct in this paper, distinguishing concepts inherited from the prior papers, concepts extended to the fleet setting, and constructs that are entirely new in Paper~6.

\begin{table}[pos=tbp]
\centering
\caption{Novelty provenance of Paper~6 constructs.}
\label{tab:novelty}
\resizebox{\columnwidth}{!}{%
\begin{tabular}{@{}lp{5.5cm}@{}}
\toprule
\textbf{Provenance} & \textbf{Constructs} \\
\midrule
\emph{Inherited} (Papers~1--5) & ECM definition, capability contracts, per-robot policy scope, single-robot recovery levels, human oversight model ($H_i$) \\
\addlinespace
\emph{Extended} to fleet & ECM registry (from local to shared $\Gamma$), recovery hierarchy (from local to fleet escalation), policy composition (from single-robot to cross-robot $\Pi$) \\
\addlinespace
\emph{New} in Paper~6 & Federation layer $\Phi$, trust-scoped delegation ($\Delta$), inter-robot capability request protocol, authority delegation semantics, fleet-level human supervision ($H_F$), coordination invariants I1--I5, architectural invariants A1--A5 \\
\bottomrule
\end{tabular}}
\end{table}


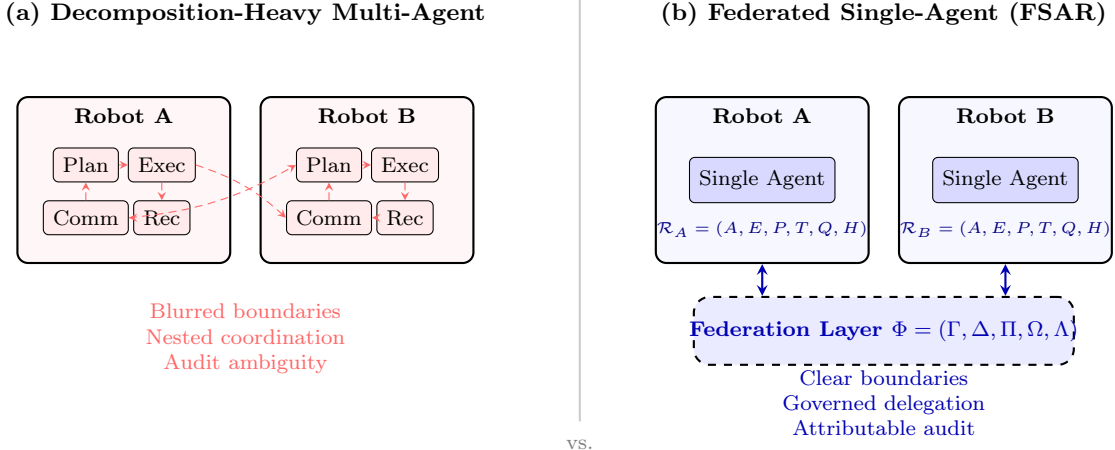
\begin{figure*}[pos=tbp]
\centering
\begin{tikzpicture}[
  robot/.style={draw, rounded corners=4pt, minimum width=2.8cm, minimum height=2.2cm, thick},
  subagent/.style={draw, fill=red!8, rounded corners=2pt, font=\scriptsize, minimum width=0.7cm, minimum height=0.45cm},
  singleagent/.style={draw, fill=blue!12, rounded corners=2pt, font=\scriptsize, minimum width=1.8cm, minimum height=0.6cm},
  fleetlayer/.style={draw, dashed, fill=blue!5, rounded corners=6pt, thick},
  arrow/.style={->, >=stealth, thick},
  messyarrow/.style={->, >=stealth, thin, red!60, densely dashed},
  cleanarrow/.style={<->, >=stealth, blue!70!black, thick},
  labelstyle/.style={font=\footnotesize\bfseries},
]

\node[labelstyle] at (-4.2, 3.4) {(a) Decomposition-Heavy Multi-Agent};

\node[robot, fill=red!3] (r1) at (-5.8, 1.2) {};
\node[font=\scriptsize\bfseries] at (-5.8, 2.05) {Robot A};
\node[subagent] (r1a) at (-6.3, 1.4) {Plan};
\node[subagent] (r1b) at (-5.3, 1.4) {Exec};
\node[subagent] (r1c) at (-6.3, 0.7) {Comm};
\node[subagent] (r1d) at (-5.3, 0.7) {Rec};
\draw[messyarrow] (r1a) -- (r1b);
\draw[messyarrow] (r1b) -- (r1d);
\draw[messyarrow] (r1c) -- (r1a);
\draw[messyarrow] (r1d) -- (r1c);

\node[robot, fill=red!3] (r2) at (-2.6, 1.2) {};
\node[font=\scriptsize\bfseries] at (-2.6, 2.05) {Robot B};
\node[subagent] (r2a) at (-3.1, 1.4) {Plan};
\node[subagent] (r2b) at (-2.1, 1.4) {Exec};
\node[subagent] (r2c) at (-3.1, 0.7) {Comm};
\node[subagent] (r2d) at (-2.1, 0.7) {Rec};
\draw[messyarrow] (r2a) -- (r2b);
\draw[messyarrow] (r2b) -- (r2d);
\draw[messyarrow] (r2c) -- (r2a);
\draw[messyarrow] (r2d) -- (r2c);

\draw[messyarrow, bend left=12] (r1b.east) to (r2c.west);
\draw[messyarrow, bend right=12] (r1c.east) to (r2a.west);

\node[font=\scriptsize, red!60, align=center] at (-4.2, -0.9) {Blurred boundaries\\Nested coordination\\Audit ambiguity};

\node[labelstyle] at (4.2, 3.4) {(b) Federated Single-Agent (FSAR)};

\node[robot, fill=blue!3] (fr1) at (2.6, 1.2) {};
\node[font=\scriptsize\bfseries] at (2.6, 2.05) {Robot A};
\node[singleagent, fill=blue!15] at (2.6, 1.2) {Single Agent};
\node[font=\tiny, blue!50!black] at (2.6, 0.55) {$\mathcal{R}_A = (A, E, P, T, Q, H)$};

\node[robot, fill=blue!3] (fr2) at (5.8, 1.2) {};
\node[font=\scriptsize\bfseries] at (5.8, 2.05) {Robot B};
\node[singleagent, fill=blue!15] at (5.8, 1.2) {Single Agent};
\node[font=\tiny, blue!50!black] at (5.8, 0.55) {$\mathcal{R}_B = (A, E, P, T, Q, H)$};

\node[fleetlayer, minimum width=5.0cm, minimum height=0.9cm, fill=blue!8] (fed) at (4.2, -0.8) {};
\node[font=\scriptsize\bfseries, blue!70!black] at (4.2, -0.8) {Federation Layer $\Phi = (\Gamma, \Delta, \Pi, \Omega, \Lambda)$};

\draw[cleanarrow] (fr1.south) -- (fr1.south |- fed.north);
\draw[cleanarrow] (fr2.south) -- (fr2.south |- fed.north);

\node[font=\scriptsize, blue!70!black, align=center] at (4.2, -1.75) {Clear boundaries\\Governed delegation\\Attributable audit};

\draw[thick, gray!40] (0.2, 3.6) -- (0.2, -2.0);
\node[font=\scriptsize, gray] at (0.2, -2.3) {vs.};

\end{tikzpicture}
\caption{Decomposition-heavy multi-agent architecture (a) versus federated single-agent architecture (b). In~(a), each robot is internally fragmented into multiple fleet-visible agents, producing nested cross-agent coordination with blurred boundaries. In~(b), each robot remains a single coherent agent; coordination occurs through a governed federation layer.}
\label{fig:hero}
\end{figure*}

\section{Motivation: From Single-Agent Robots to Federated Fleets}
\label{sec:motivation}

\subsection{Single-Agent Runtime Principle}
\label{sec:motivation:recap}

Papers~1--5 of this series established that one robot constitutes one persistent embodied agent, with a local runtime identity, installed capability modules (ECMs), policy scope, recovery authority, and human supervision interface~\cite{qin2025paper1,qin2025paper2,qin2025paper3,qin2025paper4,qin2025paper5}. Internal software components---planners, controllers, perception modules---are implementation mechanisms, not independent principals. This principle avoids the governance costs that arise when every internal component is modeled as a separate agent: ambiguous failure ownership, unclear policy authority, and fragmented human oversight.

\subsection{Why Multi-Robot Does Not Imply Multi-Agent Inside Every Robot}
\label{sec:motivation:why-not}

When the problem shifts from one robot to many robots, a common intuition is that multi-robot coordination should drive increased agent decomposition at every level~\cite{wooldridge2009introduction,jennings1998roadmap}. If robots must coordinate, then perhaps each robot should also be decomposed into multiple agents---a planning agent, a communication agent, a coordination agent, a safety agent---so that inter-robot coordination can be expressed as inter-agent coordination at all levels.

This intuition is understandable but carries significant systems costs. Intra-robot multi-agent fragmentation in the fleet context introduces five specific problems: (1)~\emph{responsibility boundary confusion}---failures may be attributable to any internal agent within any robot, making localization difficult; (2)~\emph{recovery ownership ambiguity}---when recovery is distributed across internal agents, the question of who initiates recovery becomes an internal coordination problem that must be solved before fleet-level recovery can be invoked; (3)~\emph{policy scope fragmentation}---cross-robot policy composition must account for both inter-robot and intra-robot policy inconsistencies; (4)~\emph{audit trail complexity}---each robot presenting multiple principals multiplies the dimensions of attribution; and (5)~\emph{human oversight fragmentation}---supervisors who think in terms of ``Robot~B opened the door'' must instead trace through nested internal agent attribution chains.

These problems arise from the structural decision to multiply fleet-visible agent boundaries. FSAR avoids them by preserving the single-agent boundary and pushing coordination to the inter-robot layer. We therefore adopt a different organizational principle:

\begin{quote}
\emph{The basic coordination unit of an embodied fleet is the robot, treated as a single coherent agent. The fleet is a federation of such agents, not a flat collection of sub-agents drawn from multiple robots.}
\end{quote}

This implies that inter-robot coordination is expressed as relations between robot-level agents; each robot's internal architecture is invisible to the fleet; and fleet coordination---task allocation, capability sharing, failure recovery, trust management, and human oversight---is organized at the federation layer. (We use ``fleet coordination'' throughout to denote the organized management of cross-robot interactions; we avoid the term ``fleet intelligence'' to prevent confusion with emergent or centralized decision-making.)

\section{FSAR Model and Coordination Semantics}
\label{sec:model}

\subsection{Robot as a Single Embodied Agent}
\label{sec:model:robot}

We model each robot $r_i$ as a single embodied agent with a persistent local runtime. The runtime of robot $r_i$ is represented as:
\begin{equation}
\label{eq:runtime}
\mathcal{R}_i = (A_i,\; E_i,\; P_i,\; T_i,\; Q_i,\; H_i)
\end{equation}
where $A_i$ denotes the single embodied agent identity of robot $r_i$; $E_i$ denotes the locally installed embodied capability set (ECMs); $P_i$ denotes the local policy scope governing permissible actions; $T_i$ denotes the trust state through which external robots may interact with $r_i$; $Q_i$ denotes the local recovery authority and escalation policy; and $H_i$ denotes the local human supervision interface, if any.

\paragraph{Embodied Capability Modules (ECMs).} An ECM is the unit of executable capability within a robot runtime. Each ECM $e \in E_i$ encapsulates a specific embodied capability (e.g., ``navigate.indoor,'' ``carry.package,'' ``door.open.secure'') together with its contract---preconditions, postconditions, resource requirements, and version metadata. ECMs are introduced in our prior work~\cite{qin2025paper1,qin2025paper5} and generalized here to the fleet setting. This paper is self-contained: ECMs can be understood as typed, versioned capability units that a robot may possess, advertise, execute, or delegate.

\paragraph{Capability surface.} The \emph{capability surface} of robot $r_i$ is the set of ECMs that $r_i$ currently advertises as available for external coordination: $\mathrm{Surface}(r_i) \subseteq E_i$. Not all locally installed ECMs are necessarily advertised---a robot may possess a capability without exposing it to the fleet, depending on trust, policy, and availability constraints. The capability surface is the fleet-visible interface of the robot, analogous to a public API in software engineering.

The central design commitment is that $A_i$ is \emph{singleton} at the robot-runtime level: the robot is treated as one coherent embodied agent even if its internal implementation contains multiple software components, models, controllers, or services. These internal elements do not constitute independent agent principals for fleet coordination. Note that the term ``agent'' in FSAR refers to the robot as a whole---a persistent embodied runtime with physical identity and local authority---rather than to fine-grained software agents in the MAS tradition~\cite{wooldridge2009introduction}.

\subsection{Fleet as a Federation of Single-Agent Runtimes}
\label{sec:model:fleet}

A fleet of $n$ robots is modeled as a federation:
\begin{equation}
\label{eq:fleet}
\mathcal{F} = \{\mathcal{R}_1,\; \mathcal{R}_2,\; \dots,\; \mathcal{R}_n\}
\end{equation}
augmented with a federation layer:
\begin{equation}
\label{eq:federation}
\Phi = (\Gamma,\; \Delta,\; \Pi,\; \Omega,\; \Lambda)
\end{equation}
where $\Gamma$ is the shared ECM registry and capability advertisement layer; $\Delta$ is the cross-robot delegation and coordination protocol; $\Pi$ is the fleet-level policy composition layer; $\Omega$ is the fleet-level recovery orchestration layer; and $\Lambda$ is the hierarchical human oversight layer.

The full federated fleet model is therefore:
\begin{equation}
\label{eq:fleet-model}
\mathcal{M}_{\text{fleet}} = (\mathcal{F},\; \Phi)
\end{equation}

This representation emphasizes that fleet coordination is not stored as a global super-agent that absorbs all local runtimes, nor as a nested society of internal robot agents. Instead, coordination emerges from relations defined across coherent local runtimes.


\begin{figure*}[pos=tbp]
\centering
\begin{tikzpicture}[
  localbox/.style={draw, rounded corners=3pt, fill=green!5, minimum width=2.4cm, minimum height=3.6cm, thick},
  component/.style={draw, rounded corners=2pt, fill=white, font=\scriptsize, minimum width=2cm, minimum height=0.4cm, align=center},
  fedcomp/.style={draw, rounded corners=2pt, fill=blue!10, font=\scriptsize, minimum width=2.1cm, minimum height=0.4cm, align=center},
  oversight/.style={draw, rounded corners=2pt, fill=orange!10, font=\scriptsize, minimum width=2.1cm, minimum height=0.4cm, align=center},
  layerlabel/.style={font=\footnotesize\bfseries, rotate=90},
  arrow/.style={->, >=stealth, thick, blue!60},
]

\node[layerlabel, green!50!black] at (-6.2, 0) {Local Runtimes};
\node[layerlabel, blue!60!black] at (-6.2, 4.2) {Federation Layer};
\node[layerlabel, orange!60!black] at (-6.2, 6.5) {Oversight};

\foreach \x/\name/\idx in {-4/Robot A/1, -1/Robot B/2, 2/Robot C/3, 5/Robot D/4} {
  \node[localbox] (lr\idx) at (\x, 0) {};
  \node[font=\scriptsize\bfseries, green!40!black] at (\x, 1.55) {\name};
  \node[component] at (\x, 1.1) {Agent Core $A_{\idx}$};
  \node[component] at (\x, 0.5) {ECM Mgr $E_{\idx}$};
  \node[component] at (\x, -0.1) {Policy $P_{\idx}$};
  \node[component] at (\x, -0.7) {Recovery $Q_{\idx}$};
  \node[component] at (\x, -1.3) {Human $H_{\idx}$};
}

\draw[thick, blue!40, dashed, rounded corners=6pt, fill=blue!3] (-5.3, 3) rectangle (6.3, 5.4);
\node[font=\footnotesize\bfseries, blue!60!black] at (0.5, 5.1) {Fleet Federation Layer $\Phi$};

\node[fedcomp] (gamma) at (-3.5, 4.4) {Registry $\Gamma$};
\node[fedcomp] (delta) at (-0.7, 4.4) {Coord. Engine $\Delta$};
\node[fedcomp] (pi) at (2.1, 4.4) {Policy Resolver $\Pi$};
\node[fedcomp] (omega) at (-2.1, 3.5) {Recovery Orch. $\Omega$};
\node[fedcomp] (xi) at (0.9, 3.5) {Audit $\Xi$};
\node[fedcomp] (trust) at (3.9, 3.5) {Trust \& Auth Mgr};
\node[fedcomp] (lambda) at (4.9, 4.4) {Oversight $\Lambda$};

\draw[thick, orange!40, dashed, rounded corners=6pt, fill=orange!3] (-5.3, 5.8) rectangle (6.3, 7.1);
\node[font=\footnotesize\bfseries, orange!60!black] at (0.5, 6.8) {Fleet Human Oversight};
\node[oversight] (hf) at (-1.5, 6.3) {Fleet Supervisor $H_F$};
\node[oversight] (mon) at (2.5, 6.3) {Audit / Monitoring};

\foreach \idx in {1,2,3,4} {
  \draw[arrow, thin] (lr\idx.north) -- ++(0, 0.85);
}

\draw[arrow] (lambda.north) -- ++(0, 0.55);
\draw[arrow] (xi) -- (xi |- mon) -- (mon);

\end{tikzpicture}
\caption{FSAR system model: three-layer architecture. Bottom: local robot runtimes, each containing an agent core, ECM manager, policy engine, recovery manager, and human interface. Middle: fleet federation layer with shared registry, coordination engine, policy resolver, recovery orchestrator, audit service, and trust manager. Top: fleet-level human oversight.}
\label{fig:model}
\end{figure*}
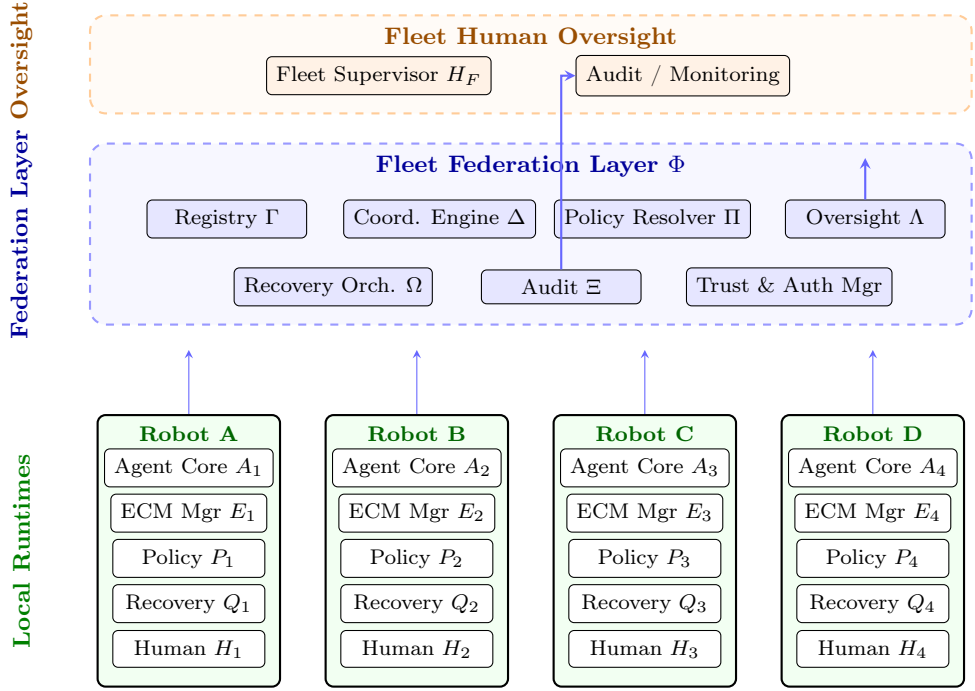

\subsection{Local Capability Sets and Federated Discoverability}
\label{sec:model:capability}

Each robot $r_i$ maintains a local capability set $E_i = \{e_{i1}, e_{i2}, \dots, e_{ik}\}$, where each $e_{ij}$ is a locally executable ECM with its own contract, version state, and policy requirements.

We distinguish three layers of capability presence:
\begin{itemize}[leftmargin=*]
  \item \textbf{Local capability possession}: $e \in E_i$.
  \item \textbf{Federated capability advertisement}: the capability is visible through the shared registry $\Gamma$.
  \item \textbf{Delegable capability availability}: another robot may request the capability under applicable trust, policy, and authority constraints.
\end{itemize}

Formally:
\begin{align}
\Possess(r_i, e) &\iff e \in E_i \label{eq:possess} \\
\Advertise(r_i, e) &\iff \Possess(r_i, e) \nonumber\\
  &\quad \wedge\; \mathrm{Visible}_\Gamma(r_i, e) \label{eq:advertise} \\
\Delegable(r_i, e, r_j) &\iff \Advertise(r_i, e) \nonumber\\
  &\quad \wedge\; \mathrm{TrustOK}(r_j, r_i, e) \nonumber\\
  &\quad \wedge\; \mathrm{PolicyOK}(r_j, r_i, e) \label{eq:delegable}
\end{align}

A robot may possess a capability without advertising it to the fleet, and may advertise it without permitting arbitrary cross-robot delegation.

\subsection{Inter-Robot Capability Requests}
\label{sec:model:request}

A central operation in FSAR is the inter-robot capability request. Suppose robot $r_i$ requires capability $e$, which it does not locally possess or cannot currently execute. It may request robot $r_j$ to execute $e$ on its behalf:
\begin{equation}
\label{eq:request}
\Req(r_i,\; r_j,\; e,\; \sigma)
\end{equation}
where $\sigma$ denotes the task context or request scope.

A request is admissible if and only if:
\begin{equation}
\label{eq:admissible}
\begin{split}
&\mathrm{AdmissibleReq}(r_i, r_j, e, \sigma) \iff \\
&\quad \Possess(r_j, e) \;\wedge\; \Advertise(r_j, e) \\
&\quad \wedge\; \mathrm{TrustOK}(r_i, r_j, e, \sigma) \\
&\quad \wedge\; \mathrm{PolicyComposeOK}(r_i, r_j, e, \sigma) \\
&\quad \wedge\; \mathrm{LocalExecPreserved}(r_j, e)
\end{split}
\end{equation}

The final condition is especially important: \emph{delegation is not transfer of agenthood}. Robot~$r_j$ executes the requested capability as its own local agent, under its own runtime, policies, and recovery authority.

\paragraph{Request lifecycle.} An inter-robot capability request proceeds through five phases:
\begin{enumerate}[leftmargin=*, nosep, label=(\arabic*)]
  \item \emph{Gap detection}: $r_i$ determines $e \notin E_i$ or $\neg\mathrm{Executable}(r_i, e, \sigma)$.
  \item \emph{Federated discovery}: $r_i$ queries $\Gamma$ for candidates $\{r_j \mid \Advertise(r_j, e) \wedge r_j \neq r_i\}$.
  \item \emph{Request formulation}: $r_i$ constructs $\Req(r_i, r_j, e, \sigma)$.
  \item \emph{Local evaluation}: $r_j$ performs admissibility checking, yielding one of \texttt{accept}, \texttt{defer}, \texttt{negotiate}, or \texttt{reject}.
  \item \emph{Execution and completion}: $r_j$ executes $e$ locally and returns
    $\Result(r_j, e, \sigma) \in \{\texttt{success}(\cdot),$ $\texttt{partial}(\cdot), \texttt{failure}(\cdot)\}$.
\end{enumerate}

\paragraph{Boundary preservation.} Inter-robot requests operate at the capability surface, not the internal runtime surface. The requesting robot specifies \emph{what} it needs, not \emph{how} the executing robot should achieve it. This yields three guarantees: no remote planning injection ($r_i$ cannot dictate $r_j$'s internal strategy); no remote policy override ($r_i$'s request does not relax $P_j$); and no remote recovery takeover (failure during execution is first handled by $Q_j$).


\begin{figure}[pos=tbp]
\centering
\begin{tikzpicture}[
  phase/.style={draw, rounded corners=3pt, fill=blue!8, font=\scriptsize, minimum width=2.6cm, minimum height=0.6cm, align=center, thick},
  arrow/.style={->, >=stealth, thick, blue!60!black},
  note/.style={font=\tiny, gray!70!black, align=left},
]

\node[phase] (p1) at (0, 4.5) {1. Gap Detection};
\node[phase] (p2) at (0, 3.3) {2. Federated Discovery};
\node[phase] (p3) at (0, 2.1) {3. Request Formulation};
\node[phase] (p4) at (0, 0.9) {4. Local Evaluation};
\node[phase] (p5) at (0, -0.3) {5. Execution \& Result};

\draw[arrow] (p1) -- (p2);
\draw[arrow] (p2) -- (p3);
\draw[arrow] (p3) -- (p4);
\draw[arrow] (p4) -- (p5);

\node[note, right] at (1.6, 4.5) {$e \notin E_i$ or not executable};
\node[note, right] at (1.6, 3.3) {Query $\Gamma$ for candidates};
\node[note, right] at (1.6, 2.1) {$\Req(r_i, r_j, e, \sigma)$};
\node[note, right] at (1.6, 0.9) {accept / defer / negotiate / reject};
\node[note, right] at (1.6, -0.3) {$r_j$ executes under $\mathcal{R}_j$};

\node[font=\tiny\bfseries, blue!50!black, rotate=90] at (-2.2, 3.9) {Requester $r_i$};
\node[font=\tiny\bfseries, green!50!black, rotate=90] at (-2.2, 0.3) {Executor $r_j$};
\draw[thick, gray!30] (-2.0, 2.7) -- (-1.5, 2.7);

\end{tikzpicture}
\caption{Five-phase lifecycle of an inter-robot capability request. Phases~1--3 are driven by the requesting robot $r_i$; phases~4--5 are driven by the executing robot $r_j$ under its own local runtime.}
\label{fig:lifecycle}
\end{figure}
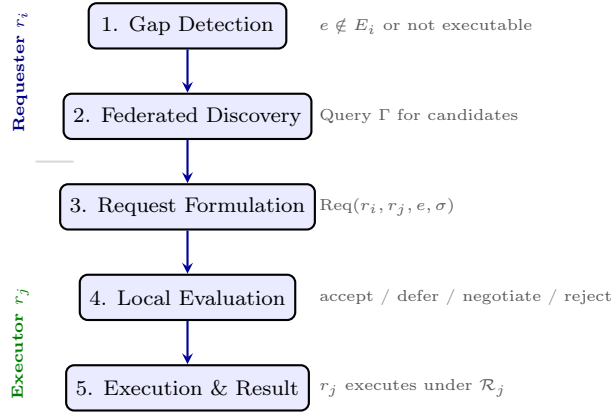

\subsection{Authority Layers}
\label{sec:model:authority}

For a request $\Req(r_i, r_j, e, \sigma)$, we define four authority dimensions:
\begin{itemize}[leftmargin=*]
  \item \textbf{Request authority} ($a^{\mathrm{req}}$): whether $r_i$ may issue the request.
  \item \textbf{Execution authority} ($a^{\mathrm{exec}}$): whether $r_j$ may execute $e$ under the request.
  \item \textbf{Override authority} ($a^{\mathrm{ovr}}$): whether the request may override local scheduling or conflict resolution.
  \item \textbf{Audit authority} ($a^{\mathrm{audit}}$): whether the request and execution must be recorded under local, fleet, or human-visible audit trails.
\end{itemize}

We write the authority tuple as:
\begin{equation}
\label{eq:auth}
\Auth(r_i, r_j, e, \sigma) = (a^{\mathrm{req}},\; a^{\mathrm{exec}},\; a^{\mathrm{ovr}},\; a^{\mathrm{audit}})
\end{equation}

This layered view prevents conflating the right to ask with the right to command, and conflating the right to execute with the right to override local runtime decisions.

\paragraph{Request and execution authority.} Request authority ($a^{\mathrm{req}}$) is governed by trust relations, fleet-level role assignments, and task-specific delegation chains. Execution authority ($a^{\mathrm{exec}}$) is local to $r_j$ and depends on $P_j$, the current runtime state, and ECM contract constraints. A critical principle is that \emph{execution authority is never automatically implied by request authority}: the fact that $r_i$ is authorized to ask does not mean $r_j$ is obligated to comply.

\paragraph{Override authority} determines whether a fleet-level request may override local scheduling. FSAR treats override authority as exceptional and scoped:
\begin{equation}
\label{eq:override}
a^{\mathrm{ovr}} \in \{\texttt{none},\; \texttt{soft\text{-}priority},\; \texttt{hard\text{-}preempt},\; \texttt{emergency\text{-}only}\}
\end{equation}

\paragraph{Non-transitivity of delegation chains.} When a fleet task involves a chain of delegations---$r_i$ requests $r_j$, and $r_j$ in turn requests $r_k$---authority does not propagate transitively by default:
\begin{equation}
\label{eq:nontransitive}
\begin{split}
a^{\mathrm{req}}(r_i, r_j, e_1, \sigma) &\;\wedge\; a^{\mathrm{req}}(r_j, r_k, e_2, \sigma) \\
&\;\not\Rightarrow\; a^{\mathrm{req}}(r_i, r_k, e_2, \sigma)
\end{split}
\end{equation}
A delegation chain $\Chain(\sigma) = [(\Req_1, \Auth_1), \dots, (\Req_m, \Auth_m)]$ is \emph{well-formed} if and only if every link is independently admissible.

\paragraph{Authority revocation.} Authority in FSAR is not permanent. Revocation may be triggered by trust downgrade (repeated failures or policy violations), context expiration (task scope $\sigma$ has concluded), or human override. Revocation takes effect at the next request boundary---it does not interrupt an already-executing capability.


\begin{figure}[pos=tbp]
\centering
\begin{tikzpicture}[
  dim/.style={draw, rounded corners=3pt, thick, minimum width=2.8cm, minimum height=0.55cm, font=\scriptsize, align=center},
  arrow/.style={->, >=stealth, thick},
]

\node[dim, fill=red!8, anchor=west] (req) at (-0.6, 3.2) {$a^{\mathrm{req}}$: Request Authority};
\node[dim, fill=green!8, anchor=west] (exec) at (-0.6, 2.3) {$a^{\mathrm{exec}}$: Execution Authority};
\node[dim, fill=orange!8, anchor=west] (ovr) at (-0.6, 1.4) {$a^{\mathrm{ovr}}$: Override Authority};
\node[dim, fill=blue!8, anchor=west] (aud) at (-0.6, 0.5) {$a^{\mathrm{audit}}$: Audit Authority};

\draw[thick, decorate, decoration={brace, amplitude=5pt, mirror}] (-1.0, 0.1) -- (-1.0, 3.55);
\node[font=\scriptsize\bfseries, rotate=90] at (-1.6, 1.85) {$\Auth(r_i, r_j, e, \sigma)$};

\node[font=\scriptsize\bfseries] at (5, 3.9) {Delegation Chain};

\node[draw, circle, fill=blue!10, minimum size=0.6cm, font=\scriptsize\bfseries] (ri) at (5, 3.2) {$r_i$};
\node[draw, circle, fill=blue!10, minimum size=0.6cm, font=\scriptsize\bfseries] (rj) at (5, 1.85) {$r_j$};
\node[draw, circle, fill=blue!10, minimum size=0.6cm, font=\scriptsize\bfseries] (rk) at (5, 0.5) {$r_k$};

\draw[arrow, thick] (ri) -- node[left, font=\tiny] {$\Auth_1$} (rj);
\draw[arrow, thick] (rj) -- node[left, font=\tiny] {$\Auth_2$} (rk);

\draw[arrow, dashed, red!60, thick] (ri.east) to[out=-20, in=20, looseness=1.3] node[right, font=\tiny, red!60, pos=0.5] {$\not\Rightarrow$ transitive} (rk.east);

\node[font=\tiny, gray] at (5, -0.2) {Non-transitive by default};

\end{tikzpicture}
\caption{Left: the four-dimensional authority tuple. Right: delegation chain non-transitivity---authority from $r_i$ to $r_j$ and $r_j$ to $r_k$ does not imply authority from $r_i$ to $r_k$.}
\label{fig:authority}
\end{figure}
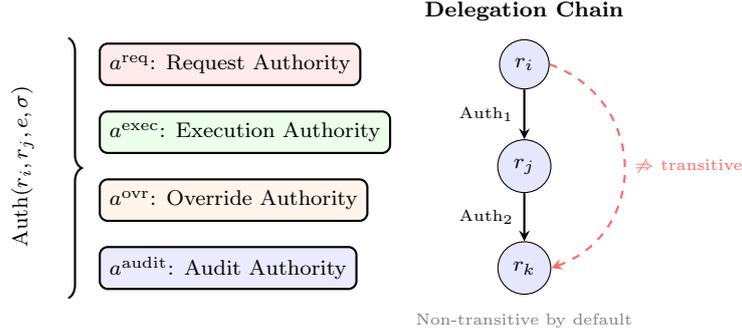

\subsection{Trust Scopes}
\label{sec:model:trust}

Trust in federated fleets is scope-bound:
\begin{multline}
\label{eq:trust}
\Trust(r_i, r_j, e, \sigma) \in {} \\
\{\texttt{none},\; \texttt{capability},\; \texttt{task},\; \texttt{session},\; \texttt{persistent}\}
\end{multline}

These levels range from no delegation (\texttt{none}) to long-lived trusted peer relations (\texttt{persistent}). A robot may trust another robot for door opening in one task but not for access to sensitive inspection or close human-interaction capabilities. Trust is not static: it may evolve based on performance history, failures, operator approvals, or policy changes~\cite{sandholm1999distributed}. FSAR does not require a fixed trust matrix; it requires that trust be explicit, scope-aware, and queryable.

\subsection{Policy Composition Across Robots}
\label{sec:model:policy}

Each robot $r_i$ has a local policy scope $P_i$, but fleet tasks often span multiple robots. We define a policy composition function:
\begin{equation}
\label{eq:policy}
\Pi(P_i, P_j, e, \sigma) \;\rightarrow\; \{\texttt{allow},\; \texttt{deny},\; \texttt{review}\}
\end{equation}

The three-valued outcome is intentional: some interactions are not outright forbidden but require higher-level review due to context ambiguity or unusual delegation patterns. A fleet-level task is admissible only if all required local policy scopes and composed fleet-level constraints are satisfied.

\subsection{Layered Recovery}
\label{sec:model:recovery}

Recovery proceeds through a monotone escalation hierarchy:
\begin{equation}
\label{eq:recovery}
\RecLevel(r_i, \sigma) \in \{\texttt{local},\; \texttt{peer},\; \texttt{fleet},\; \texttt{human}\}
\end{equation}
with ordering $\texttt{local} \prec \texttt{peer} \prec \texttt{fleet} \prec \texttt{human}$. Recovery ownership begins locally and expands outward only when the local runtime cannot contain the failure~\cite{visinsky1994survey,carlson2005ugv}.

\paragraph{Level~1: Local recovery.} Robot $r_i$ attempts to resolve the issue within its own runtime, bounded by a recovery budget $(t_{\max}, n_{\max})$---maximum time and retry attempts before escalation.

\paragraph{Level~2: Peer-assisted recovery.} If local recovery is insufficient, $r_i$ may request assistance from a peer $r_j$, either through capability substitution or environmental assistance, subject to the same trust and authority constraints as routine coordination.

\paragraph{Level~3: Fleet reassignment.} The recovery orchestrator $\Omega$ may reassign the task to a different robot. Reassignment requires that tasks define \emph{reassignment checkpoints}---well-defined intermediate states from which another robot can resume.

\paragraph{Level~4: Human escalation.} The failure is escalated to human supervision through $\Lambda$, directed to local or fleet supervision depending on scope. Recovery actions are subject to the same authority constraints as routine coordination---a failure does not create new authority that did not exist before the failure.


\begin{figure}[pos=tbp]
\centering
\begin{tikzpicture}[
  level/.style={draw, rounded corners=4pt, thick, minimum width=4.5cm, minimum height=0.65cm, font=\scriptsize\bfseries, align=center},
  escalarrow/.style={->, >=stealth, thick, red!60},
  note/.style={font=\tiny, gray!60!black, align=left},
]

\node[level, fill=green!12] (l1) at (0, 3.0) {Level 1: Local Recovery};
\node[level, fill=yellow!15] (l2) at (0, 2.0) {Level 2: Peer-Assisted Recovery};
\node[level, fill=orange!12] (l3) at (0, 1.0) {Level 3: Fleet Reassignment};
\node[level, fill=red!10] (l4) at (0, 0.0) {Level 4: Human Escalation};

\draw[escalarrow] (l1.south) -- node[right, font=\tiny, xshift=2pt] {budget exceeded} (l2.north);
\draw[escalarrow] (l2.south) -- node[right, font=\tiny, xshift=2pt] {no capable peer} (l3.north);
\draw[escalarrow] (l3.south) -- node[right, font=\tiny, xshift=2pt] {no feasible reassign} (l4.north);

\node[note, anchor=west] at (2.8, 3.0) {Retry, rollback, replan\\$(t_{\max}, n_{\max})$ bounded};
\node[note, anchor=west] at (2.8, 2.0) {Capability substitution\\Environmental assist};
\node[note, anchor=west] at (2.8, 1.0) {Checkpoint-based\\Task state transfer};
\node[note, anchor=west] at (2.8, 0.0) {$H_i$ (local) or\\$H_F$ (fleet-level)};

\draw[thick, green!50!black, decorate, decoration={brace, amplitude=5pt}] (-2.8, -0.3) -- (-2.8, 3.3);
\node[font=\tiny\bfseries, green!50!black, rotate=90, align=center] at (-3.4, 1.5) {Prefer minimal\\escalation};

\end{tikzpicture}
\caption{Layered recovery hierarchy with monotone escalation. Recovery begins locally and expands outward only when the current level's budget or capacity is exhausted.}
\label{fig:recovery}
\end{figure}
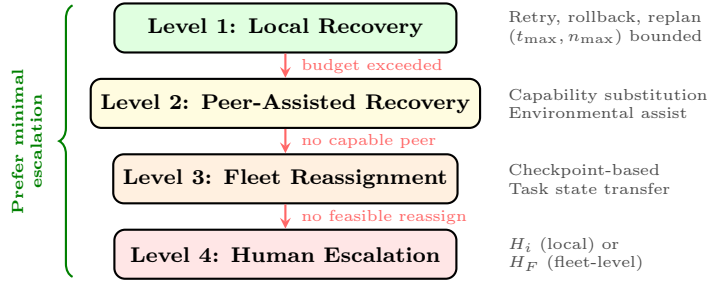

\subsection{Hierarchical Human Supervision}
\label{sec:model:supervision}

We define two primary supervisory levels: \emph{local supervision} $H_i$ (oversight specific to robot $r_i$) and \emph{fleet supervision} $H_F$ (oversight over cross-robot coordination, policy exceptions, or unresolved conflicts)~\cite{parasuraman2000model,sheridan2016humanrobot}:
\begin{equation}
\label{eq:supervision}
\Lambda = (\{H_i\}_{i=1}^{n},\; H_F)
\end{equation}
Escalation to fleet supervision occurs when a local supervisor encounters a situation involving cross-robot dependencies outside local scope. Conversely, fleet supervision does not routinely override local supervision---it intervenes only when cross-robot conflicts require fleet-level resolution. Supervision density is a policy parameter, not a structural constant; it can be adjusted at runtime based on task type, environmental conditions, and operational phase.

\subsection{Capability Advertisement}
\label{sec:model:advertisement}

Inter-robot coordination depends on a shared registry, but discoverability must be governed. Each advertisement record takes the form:
\begin{equation}
\label{eq:adv}
\begin{split}
\Adv(r_i, e) = (&\mathrm{Ver}_e,\; \mathrm{Contract}_e,\; \mathrm{TrustReq}_e,\\
  &\mathrm{PolicyReq}_e,\; \mathrm{AuthProfile}_e,\\
  &\Avail_e,\; \mathrm{Emb}_e)
\end{split}
\end{equation}
where fields encode version metadata, ECM contract, required trust scope, policy constraints, authority profile, current availability state, and embodiment context respectively. The shared registry is not a raw capability marketplace---it is a governed discovery layer. \Cref{sec:registry} develops the full registry semantics.

\subsection{Coordination Invariants}
\label{sec:model:invariants}

To preserve coherence in federated single-agent fleets, FSAR maintains five coordination invariants:

\begin{enumerate}[label=\textbf{I\arabic*},leftmargin=*]
  \item \textbf{Local agent coherence.} Each robot remains a single fleet-visible agent principal: $\forall r_i \in \mathcal{F},\; \mathrm{PrincipalCount}(r_i) = 1$.
  \item \textbf{Delegation preserves local execution ownership.}\\$\Req(r_i, r_j, e, \sigma) \Rightarrow \mathrm{LocalExecPreserved}(r_j, e)$.
  \item \textbf{Policy admissibility is jointly composed.} $\mathrm{Exec}(r_j, e, \sigma) \Rightarrow \Pi(P_i, P_j, e, \sigma) \neq \texttt{deny}$.
  \item \textbf{Recovery escalates outward monotonically.} $\texttt{local} \prec \texttt{peer} \prec \texttt{fleet} \prec \texttt{human}$.
  \item \textbf{Audit attributability.} Each cross-robot coordination event remains attributable to identifiable request, execution, and supervision principals.
\end{enumerate}

Having established the model and coordination semantics, \cref{tab:notation} summarizes all principal symbols for reference. We now turn to the infrastructure that makes them operational.

\begin{table}[pos=tbp]
\centering
\caption{Summary of principal notation.}
\label{tab:notation}
\resizebox{\columnwidth}{!}{%
\begin{tabular}{@{}cl@{}}
\toprule
\textbf{Symbol} & \textbf{Meaning} \\
\midrule
$\mathcal{R}_i$ & Local runtime of robot $r_i$: $(A_i, E_i, P_i, T_i, Q_i, H_i)$ \\
$A_i$ & Agent identity of $r_i$ \\
$E_i$ & ECM set (capabilities) of $r_i$ \\
$P_i$ & Local policy scope of $r_i$ \\
$T_i$ & Trust relations of $r_i$ \\
$Q_i$ & Recovery state of $r_i$ \\
$H_i$ & Local human supervisor for $r_i$ \\
\addlinespace
$\mathcal{F}$ & Fleet: $\{r_1, \dots, r_n\}$ \\
$\Phi$ & Federation layer: $(\Gamma, \Delta, \Pi, \Omega, \Lambda)$ \\
$\Gamma$ & Shared ECM registry \\
$\Delta$ & Cross-robot delegation protocol \\
$\Pi$ & Fleet-level policy composition \\
$\Omega$ & Fleet-level recovery orchestration \\
$\Lambda$ & Hierarchical human oversight: $(\{H_i\}, H_F)$ \\
$H_F$ & Fleet-level human supervisor \\
\addlinespace
$\Auth$ & Authority tuple: $(a^{\mathrm{req}}, a^{\mathrm{exec}}, a^{\mathrm{ovr}}, a^{\mathrm{audit}})$ \\
$\Trust$ & Trust evaluation function \\
$\Req$ & Inter-robot capability request \\
$\sigma$ & Task/coordination context \\
\bottomrule
\end{tabular}}
\end{table}

\section{Fleet Architecture and Registry Semantics}
\label{sec:architecture}
\label{sec:registry}

This section describes the runtime architecture that realizes the FSAR model and details the shared ECM registry that makes governed capability discovery operational. The architecture is organized across two layers---local robot runtimes and the fleet federation layer---with the registry as a core federation component.

\subsection{Design Goals and Architectural Overview}
\label{sec:architecture:overview}

The architecture pursues five design goals: (1)~preserve local runtime self-sufficiency; (2)~support fleet coordination without absorbing local autonomy; (3)~make every coordination event governed and auditable; (4)~support heterogeneous robots with different capabilities, policies, and trust configurations; and (5)~degrade gracefully when the federation layer is unavailable.

The \textbf{Local Runtime Layer} manages each robot's embodied capabilities, policy enforcement, recovery handling, and human supervision. The \textbf{Fleet Federation Layer} provides shared infrastructure for cross-robot coordination, including the governed capability registry. The relationship is asymmetric: the local runtime is authoritative for execution; the federation layer is authoritative for coordination. A local runtime may operate without the federation layer (isolated mode), but the federation layer cannot execute capabilities without local runtimes.


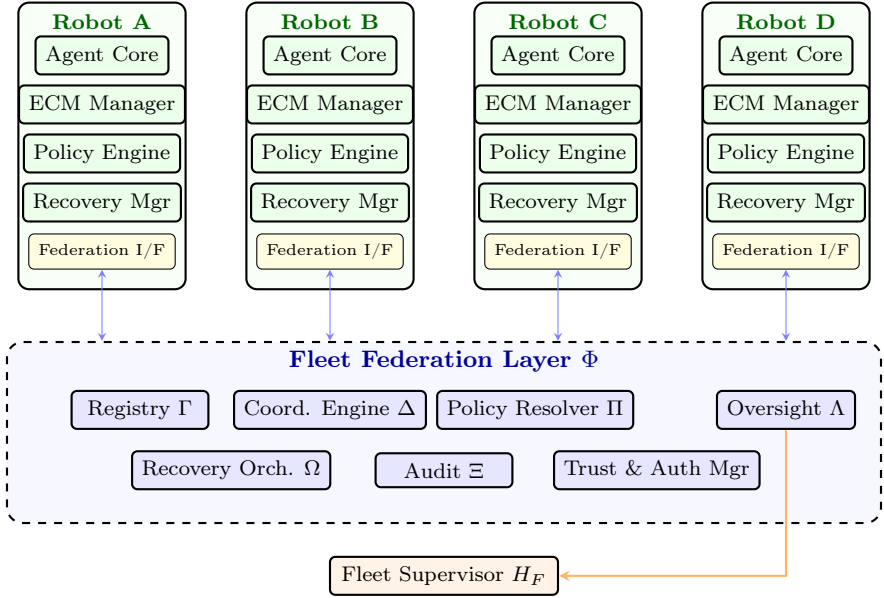
\begin{figure*}[pos=tbp]
\centering
\begin{tikzpicture}[
  localcomp/.style={draw, rounded corners=2pt, fill=green!8, font=\scriptsize, minimum width=1.6cm, minimum height=0.45cm, align=center, thick},
  fedcomp/.style={draw, rounded corners=2pt, fill=blue!10, font=\scriptsize, minimum width=1.8cm, minimum height=0.45cm, align=center, thick},
  humancomp/.style={draw, rounded corners=2pt, fill=orange!10, font=\scriptsize, minimum width=1.8cm, minimum height=0.45cm, align=center, thick},
  robotbox/.style={draw, rounded corners=4pt, fill=green!3, thick, minimum width=2.2cm},
  fedbox/.style={draw, dashed, rounded corners=6pt, fill=blue!3, thick},
  iface/.style={draw, fill=yellow!15, rounded corners=2pt, font=\tiny, minimum width=1.6cm, minimum height=0.35cm, align=center},
  arrow/.style={<->, >=stealth, thick, blue!50},
]

\node[robotbox, minimum height=3.8cm] (rA) at (-4.5, 0) {};
\node[font=\scriptsize\bfseries, green!40!black] at (-4.5, 1.65) {Robot A};
\node[localcomp] at (-4.5, 1.2) {Agent Core};
\node[localcomp] at (-4.5, 0.55) {ECM Manager};
\node[localcomp] at (-4.5, -0.1) {Policy Engine};
\node[localcomp] at (-4.5, -0.75) {Recovery Mgr};
\node[iface] (ifA) at (-4.5, -1.4) {Federation I/F};

\node[robotbox, minimum height=3.8cm] (rB) at (-1.5, 0) {};
\node[font=\scriptsize\bfseries, green!40!black] at (-1.5, 1.65) {Robot B};
\node[localcomp] at (-1.5, 1.2) {Agent Core};
\node[localcomp] at (-1.5, 0.55) {ECM Manager};
\node[localcomp] at (-1.5, -0.1) {Policy Engine};
\node[localcomp] at (-1.5, -0.75) {Recovery Mgr};
\node[iface] (ifB) at (-1.5, -1.4) {Federation I/F};

\node[robotbox, minimum height=3.8cm] (rC) at (1.5, 0) {};
\node[font=\scriptsize\bfseries, green!40!black] at (1.5, 1.65) {Robot C};
\node[localcomp] at (1.5, 1.2) {Agent Core};
\node[localcomp] at (1.5, 0.55) {ECM Manager};
\node[localcomp] at (1.5, -0.1) {Policy Engine};
\node[localcomp] at (1.5, -0.75) {Recovery Mgr};
\node[iface] (ifC) at (1.5, -1.4) {Federation I/F};

\node[robotbox, minimum height=3.8cm] (rD) at (4.5, 0) {};
\node[font=\scriptsize\bfseries, green!40!black] at (4.5, 1.65) {Robot D};
\node[localcomp] at (4.5, 1.2) {Agent Core};
\node[localcomp] at (4.5, 0.55) {ECM Manager};
\node[localcomp] at (4.5, -0.1) {Policy Engine};
\node[localcomp] at (4.5, -0.75) {Recovery Mgr};
\node[iface] (ifD) at (4.5, -1.4) {Federation I/F};

\node[fedbox, minimum width=11.5cm, minimum height=2.4cm] (fed) at (0, -3.8) {};
\node[font=\footnotesize\bfseries, blue!60!black] at (0, -2.85) {Fleet Federation Layer $\Phi$};

\node[fedcomp] (reg) at (-4, -3.5) {Registry $\Gamma$};
\node[fedcomp] (coord) at (-1.5, -3.5) {Coord. Engine $\Delta$};
\node[fedcomp] (pol) at (1.2, -3.5) {Policy Resolver $\Pi$};
\node[fedcomp] (rec) at (-2.8, -4.3) {Recovery Orch. $\Omega$};
\node[fedcomp] (aud) at (0, -4.3) {Audit $\Xi$};
\node[fedcomp] (tru) at (2.8, -4.3) {Trust \& Auth Mgr};
\node[fedcomp] (lam) at (4.5, -3.5) {Oversight $\Lambda$};

\draw[arrow, thin] (ifA.south) -- (ifA.south |- fed.north);
\draw[arrow, thin] (ifB.south) -- (ifB.south |- fed.north);
\draw[arrow, thin] (ifC.south) -- (ifC.south |- fed.north);
\draw[arrow, thin] (ifD.south) -- (ifD.south |- fed.north);

\node[humancomp, minimum width=3cm] (hf) at (0, -5.7) {Fleet Supervisor $H_F$};
\draw[->, >=stealth, thick, orange!60] (lam.south) |- (hf.east);

\end{tikzpicture}
\caption{Fleet runtime architecture. Each robot contains a self-sufficient local runtime with agent core, ECM manager, policy engine, and recovery manager, connected to the federation layer through a thin federation interface. The federation layer provides shared registry, coordination engine, policy resolver, recovery orchestrator, audit service, trust manager, and human oversight console.}
\label{fig:architecture}
\end{figure*}

\subsection{Local Robot Runtime Components}
\label{sec:architecture:local}

Each robot's local runtime $\mathcal{R}_i$ contains: an \textbf{Agent Core} (persistent identity $A_i$ and control center, the single fleet-visible principal); a \textbf{Local ECM Manager} (installing, loading, versioning, and executing ECMs in $E_i$); a \textbf{Local Policy Engine} (enforcing $P_i$ on every capability execution, request evaluation, and recovery action); a \textbf{Local Recovery Manager} (implementing $Q_i$ with local retry, rollback, replanning, and escalation); a \textbf{Federation Interface} (thin translation layer between local runtime state and the federation protocol---it publishes advertisements, evaluates incoming requests, and reports availability updates, but does not make autonomous coordination decisions); and a \textbf{Local Human Supervision Interface} ($H_i$, for monitoring, intervention, and override).

\subsection{Fleet Federation Layer Components}
\label{sec:architecture:fleet}

The federation layer $\Phi$ contains: the \textbf{Shared ECM Registry} ($\Gamma$, governed capability advertisement, structured queries, and dynamic availability tracking---detailed in \cref{sec:architecture:registry} below); the \textbf{Coordination Engine} ($\Delta$, cross-robot delegation protocol including request routing, negotiation management, delegation chain tracking, and completion tracking); the \textbf{Trust and Authority Manager} (trust state storage, authority evaluation, trust updates, and revocation); the \textbf{Fleet Policy Resolver} ($\Pi$, policy composition and fleet-level policy management---critically, it \emph{never silently overrides a local policy denial}; conflicts are escalated, not suppressed); the \textbf{Recovery Orchestrator} ($\Omega$, peer-assisted recovery, fleet reassignment, and escalation management, activated only when local recovery has failed); the \textbf{Audit and Traceability Service} ($\Xi$, recording all cross-robot coordination events with principal attribution); and the \textbf{Human Oversight Console} ($\Lambda$, fleet-level supervision interface $H_F$ with intervention controls for trust modification, request approval, and supervision density adjustment).

\subsection{Shared ECM Registry: Record Structure and Dynamics}
\label{sec:architecture:registry}

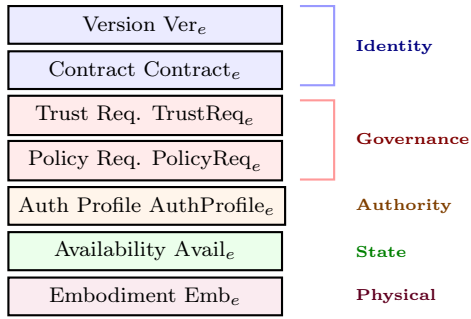
\begin{figure}[pos=tbp]
\centering
\begin{tikzpicture}[
  field/.style={draw, minimum width=3.6cm, minimum height=0.5cm, font=\scriptsize, anchor=west, thick},
  label/.style={font=\scriptsize\bfseries, anchor=east},
]

\node[font=\footnotesize\bfseries, blue!60!black] at (1.8, 4.4) {$\Adv(r_i, e)$: Advertisement Record};

\node[field, fill=blue!8] (f1) at (0, 3.7) {Version $\mathrm{Ver}_e$};
\node[field, fill=blue!8] (f2) at (0, 3.1) {Contract $\mathrm{Contract}_e$};
\node[field, fill=red!8] (f3) at (0, 2.5) {Trust Req. $\mathrm{TrustReq}_e$};
\node[field, fill=red!8] (f4) at (0, 1.9) {Policy Req. $\mathrm{PolicyReq}_e$};
\node[field, fill=orange!8] (f5) at (0, 1.3) {Auth Profile $\mathrm{AuthProfile}_e$};
\node[field, fill=green!8] (f6) at (0, 0.7) {Availability $\Avail_e$};
\node[field, fill=purple!8] (f7) at (0, 0.1) {Embodiment $\mathrm{Emb}_e$};

\draw[thick, blue!40] (3.85, 2.9) -- (4.3, 2.9) -- (4.3, 3.95) -- (3.85, 3.95);
\draw[thick, red!40] (3.85, 1.65) -- (4.3, 1.65) -- (4.3, 2.7) -- (3.85, 2.7);

\node[font=\tiny\bfseries, blue!50!black, anchor=west] at (4.45, 3.4) {Identity};
\node[font=\tiny\bfseries, red!50!black, anchor=west] at (4.45, 2.2) {Governance};
\node[font=\tiny\bfseries, orange!50!black, anchor=west] at (4.45, 1.3) {Authority};
\node[font=\tiny\bfseries, green!50!black, anchor=west] at (4.45, 0.7) {State};
\node[font=\tiny\bfseries, purple!50!black, anchor=west] at (4.45, 0.1) {Physical};

\end{tikzpicture}
\caption{Structure of a capability advertisement record in the shared ECM registry. Fields are grouped by function: identity (version, contract), governance (trust and policy requirements), authority profile, operational state (availability), and physical context (embodiment).}
\label{fig:registry}
\end{figure}

A fleet registry is not a centralized list of all installed capabilities---it is a governed discovery layer~\cite{dastani2005programming} that answers: \emph{which robot currently exposes which capability, under what contract, with what trust, policy, authority, and availability conditions?} For each advertised capability, the registry stores a governed advertisement record (\cref{eq:adv}) with seven fields: $\mathrm{Ver}_e$ (version metadata for semantic matching), $\mathrm{Contract}_e$ (ECM contract for compatibility checking), $\mathrm{TrustReq}_e$ (required trust scope for delegation), $\mathrm{PolicyReq}_e$ (policy constraints for cross-robot use), $\mathrm{AuthProfile}_e$ (authority profile), $\Avail_e$ (current availability state), and $\mathrm{Emb}_e$ (embodiment profile). The structure is shown in \cref{fig:registry}.


Availability is modeled as a first-class stateful property:
\[
\Avail(r_i, e) \in \{\texttt{ready},\; \texttt{busy},\; \texttt{degraded},\; \texttt{restricted},\; \texttt{offline}\}
\]
This richer state prevents a frequent coordination error: assuming that discoverability implies immediate utility.

The registry is authoritative about discoverability but never about execution: even after discovery succeeds, the callee robot evaluates admissibility under its local runtime. Given a request $Q = (r_A, e^*, \sigma, B, P)$, the registry returns a candidate set:
\[
\mathrm{Cand}(Q) = \{(r_j, e_j) \mid \mathrm{Match}(e_j, e^*) \wedge \Advertise(r_j, e_j)\}
\]
filtered by trust sufficiency, policy composability, authority profile, availability, and contract compatibility~\cite{meyer1992applying,qin2025paper5}. Visibility may be global, domain-scoped, or trust-gated. Registry records are dynamically updated as robots enter degraded mode, withdraw capabilities during recovery, or change advertisement during upgrade rollouts.

\subsection{Interaction Patterns and Deployment}
\label{sec:architecture:patterns}

Components interact through recurring patterns: in \emph{normal delegation}, a robot's agent core detects a capability gap, the federation interface queries the registry, the coordination engine routes the request, and the callee executes locally. In \emph{policy-flagged coordination}, the fleet policy resolver flags a request for human review. In \emph{recovery escalation}, the recovery orchestrator coordinates peer assistance or fleet reassignment. These patterns compose within a single fleet task.

The architecture supports centralized federation for small fleets, distributed federation with leader election for larger fleets, and peer-to-peer federation for intermittently connected fleets. In all topologies, the local runtime continues operating in isolated mode if the federation layer becomes unavailable.

\subsection{Architectural Invariants and Model Mapping}
\label{sec:architecture:invariants}

Five architectural invariants complement the coordination invariants I1--I5:

\begin{enumerate}[label=\textbf{A\arabic*},leftmargin=*]
  \item \textbf{Local runtime self-sufficiency.} Every robot can operate in isolated mode.
  \item \textbf{Coordination-only federation.} The federation layer never directly executes capabilities.
  \item \textbf{No silent policy override.} Policy conflicts are escalated, not suppressed.
  \item \textbf{Thin federation interface.} All execution decisions are made by local runtime components.
  \item \textbf{Audit completeness.} Every cross-robot event is recorded with principal attribution.
\end{enumerate}

\Cref{tab:mapping} provides the complete mapping from formal model elements to architectural components.

\begin{table}[pos=tbp]
\centering
\caption{Mapping from formal model elements to architectural components.}
\label{tab:mapping}
\resizebox{\columnwidth}{!}{%
\begin{tabular}{@{}ll@{}}
\toprule
\textbf{Model Element} & \textbf{Architecture Component} \\
\midrule
$A_i$ (agent identity) & Agent Core \\
$E_i$ (capability set) & Local ECM Manager \\
$P_i$ (policy scope) & Local Policy Engine \\
$T_i$ (trust state) & Trust \& Authority Manager \\
$Q_i$ (recovery authority) & Recovery Mgr + Recovery Orch. \\
$H_i$ (local supervision) & Local Human Supervision I/F \\
$\Gamma$ (shared registry) & Shared ECM Registry \\
$\Delta$ (coord.\ protocol) & Coordination Engine \\
$\Pi$ (fleet policy) & Fleet Policy Resolver \\
$\Omega$ (recovery orch.) & Recovery Orchestrator \\
$\Lambda$ (human oversight) & Human Oversight Console \\
$\Xi$ (audit) & Audit \& Traceability Service \\
\bottomrule
\end{tabular}}
\end{table}

\section{Evaluation}
\label{sec:evaluation}
\label{sec:prototype}

To validate that FSAR's abstractions compose into a working system, we conduct a simulation-based evaluation of both the coordination prototype and governance properties. A simulation-based approach is appropriate because the evaluation targets \emph{governance properties}---authority attribution, recovery escalation, policy composition---that are observable at the protocol level and do not require physical dynamics.

\subsection{Prototype and Fleet Configuration}
\label{sec:prototype:design}

The prototype instantiates all components defined in \cref{sec:architecture}: shared ECM registry, coordination engine with full request lifecycle, trust and authority evaluation, policy composition, layered recovery orchestration, audit logging, and hierarchical human oversight. The prototype fleet consists of four heterogeneous robots:
\begin{itemize}[leftmargin=*, nosep]
  \item Robot~A (Delivery): \texttt{navigate.indoor}, \texttt{carry.package}, \texttt{handover.dropoff}
  \item Robot~B (Access): \texttt{navigate.indoor}, \texttt{door.open.basic}, \texttt{door.open.secure}
  \item Robot~C (Inspection): \texttt{navigate.indoor}, \texttt{inspect.visual}, \texttt{inspect.private\_zone}
  \item Robot~D (Heavy Transport): \texttt{navigate.indoor}, \texttt{carry.heavy}, \texttt{grasp.robust}
\end{itemize}
Trust configuration: $A \leftrightarrow B$ task-scoped; $A \leftrightarrow D$ capability-scoped; $B \leftrightarrow C$ session-scoped; $C$'s \texttt{inspect.\allowbreak{}private\_zone} visible only to fleet supervisors and Robot~B; $D$ has no trust relation with~$C$.

\subsection{Coordination Workflows}
\label{sec:prototype:workflows}

\begin{figure}[pos=tbp]
\centering
\begin{tikzpicture}[
  entity/.style={draw, rounded corners=3pt, fill=#1, font=\scriptsize\bfseries, minimum width=1.2cm, minimum height=0.5cm, align=center},
  entity/.default={blue!10},
  msg/.style={->, >=stealth, thick, font=\tiny},
  timeline/.style={thick, gray!40},
  steplbl/.style={font=\tiny, fill=white, inner sep=1pt},
]

\node[entity=green!10] (rA) at (0, 5.5) {Robot A};
\node[entity=blue!10] (reg) at (2.5, 5.5) {Registry $\Gamma$};
\node[entity=green!10] (rB) at (5, 5.5) {Robot B};

\draw[timeline] (0, 5.1) -- (0, -0.5);
\draw[timeline] (2.5, 5.1) -- (2.5, -0.5);
\draw[timeline] (5, 5.1) -- (5, -0.5);

\draw[msg] (0, 4.5) -- node[steplbl, above] {1. query: door.open.secure} (2.5, 4.5);
\draw[msg] (2.5, 3.8) -- node[steplbl, above] {2. candidates: $\{r_B\}$} (0, 3.8);
\draw[msg] (0, 3.0) -- node[steplbl, above] {3. $\Req(r_A, r_B, e, \sigma)$} (5, 3.0);
\node[draw, rounded corners=2pt, fill=gray!8, font=\tiny, align=center, minimum width=1.6cm] at (5.8, 2.15) {4. evaluate\\$\Trust = \texttt{task}$\\$\Pi \to \texttt{allow}$};
\draw[msg] (5, 1.3) -- node[steplbl, above] {5. accept} (0, 1.3);
\node[font=\tiny, gray, align=center] at (5.8, 0.6) {execute under $\mathcal{R}_B$};
\draw[msg, green!50!black] (5, -0.1) -- node[steplbl, above] {6. success(door\_opened)} (0, -0.1);

\end{tikzpicture}
\caption{Sequence diagram for Workflow~1 (Door Relay). Robot~A discovers Robot~B's door-opening capability through the registry, issues a governed request, and Robot~B executes under its own local runtime.}
\label{fig:wf1}
\end{figure}
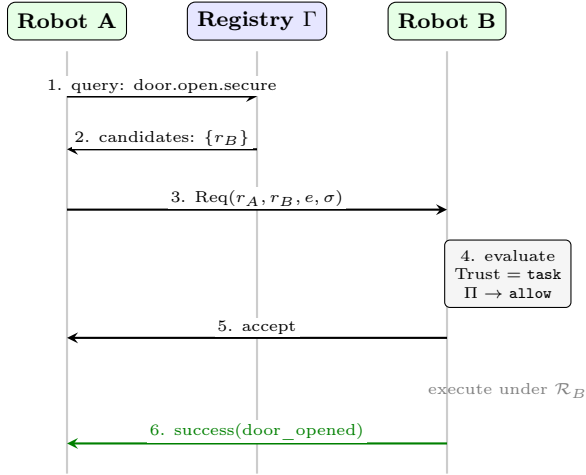

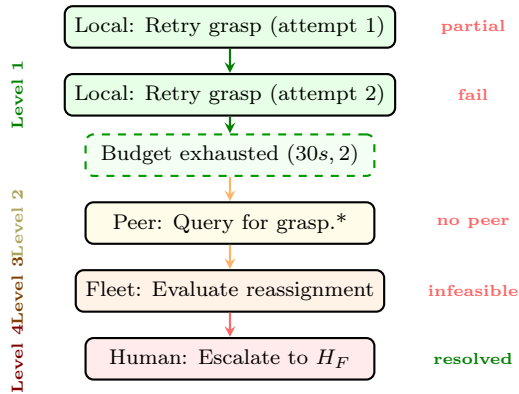
\begin{figure}[pos=tbp]
\centering
\begin{tikzpicture}[
  stepbox/.style={draw, rounded corners=3pt, thick, minimum width=3.8cm, minimum height=0.55cm, font=\scriptsize, align=center},
  result/.style={font=\tiny\bfseries, align=center},
  arrow/.style={->, >=stealth, thick},
]

\node[stepbox, fill=green!10] (s1) at (0, 5.0) {Local: Retry grasp (attempt 1)};
\node[result, red!60] at (3.2, 5.0) {partial};

\node[stepbox, fill=green!10] (s2) at (0, 4.1) {Local: Retry grasp (attempt 2)};
\node[result, red!60] at (3.2, 4.1) {fail};

\node[stepbox, fill=green!5, draw=green!60!black, dashed] (s2b) at (0, 3.3) {Budget exhausted $(30s, 2)$};

\node[stepbox, fill=yellow!10] (s3) at (0, 2.4) {Peer: Query for grasp.*};
\node[result, red!60] at (3.2, 2.4) {no peer};

\node[stepbox, fill=orange!10] (s4) at (0, 1.5) {Fleet: Evaluate reassignment};
\node[result, red!60] at (3.2, 1.5) {infeasible};

\node[stepbox, fill=red!8] (s5) at (0, 0.6) {Human: Escalate to $H_F$};
\node[result, green!50!black] at (3.2, 0.6) {resolved};

\draw[arrow, green!50!black] (s1) -- (s2);
\draw[arrow, green!50!black] (s2) -- (s2b);
\draw[arrow, orange!60] (s2b) -- (s3);
\draw[arrow, orange!60] (s3) -- (s4);
\draw[arrow, red!60] (s4) -- (s5);

\node[font=\tiny\bfseries, green!50!black, rotate=90] at (-2.8, 4.1) {Level 1};
\node[font=\tiny\bfseries, yellow!60!black, rotate=90] at (-2.8, 2.4) {Level 2};
\node[font=\tiny\bfseries, orange!50!black, rotate=90] at (-2.8, 1.5) {Level 3};
\node[font=\tiny\bfseries, red!50!black, rotate=90] at (-2.8, 0.6) {Level 4};

\end{tikzpicture}
\caption{Workflow~3: layered recovery escalation for Robot~D's grasp degradation. Recovery proceeds through all four levels---local retry, peer query, fleet reassignment evaluation, and human escalation---demonstrating monotone escalation with containment attempt at each level.}
\label{fig:wf3}
\end{figure}

Three workflows validate distinct layers of the coordination model. \textbf{Workflow~1 (Door Relay):} Robot~A detects a capability gap ($\texttt{door.open.secure} \notin E_A$), discovers Robot~B via the registry, issues a trust-evaluated request ($\Trust = \texttt{task}$, authority $(1,1,0,1)$, policy $\to$ \texttt{allow}), and Robot~B executes under its own runtime $\mathcal{R}_B$. This demonstrates structured delegation, trust scoping, policy composition, and auditability (\cref{fig:wf1}). \textbf{Workflow~2 (Layered Recovery):} Robot~D's grasp actuator degrades during heavy transport. Recovery escalates through all four levels: local retry (budget exhausted after 2 attempts / 30s), peer query (no capable peer), fleet reassignment (infeasible), and human escalation to $H_F$ (\cref{fig:wf3}). This demonstrates monotone escalation, recovery budget enforcement, and audit trails through all levels. \textbf{Workflow~3 (Human Supervision):} Robots~B and~D simultaneously request Robot~A's carrying capability; the policy resolver produces \texttt{review}, escalating to $H_F$ who prioritizes Robot~B's time-critical delivery. This demonstrates fleet-level supervision, the \texttt{review} outcome, and audited supervisor decisions. \Cref{tab:coverage} summarizes mechanism coverage across workflows.


\label{sec:prototype:wf3b}

\begin{table}[pos=tbp]
\centering
\caption{FSAR mechanisms exercised by each workflow.}
\label{tab:coverage}
\resizebox{\columnwidth}{!}{%
\begin{tabular}{@{}lccc@{}}
\toprule
\textbf{Mechanism} & \textbf{WF1} & \textbf{WF2} & \textbf{WF3} \\
\midrule
Inter-robot request & $\bullet$ & via recovery & $\bullet$ \\
Trust evaluation & task & --- & task \\
Authority tuple & standard & --- & standard \\
Policy composition & allow & --- & review \\
Registry discovery & single & no peer & conflict \\
Layered recovery & --- & all 4 levels & --- \\
Fleet reassignment & --- & attempted & --- \\
Human escalation & --- & Level~4 & conflict \\
Delegation chain & single & --- & --- \\
\bottomrule
\end{tabular}}
\end{table}

\subsection{Evaluation Goals}
\label{sec:evaluation:goals}

The evaluation is designed to answer five questions:
\begin{enumerate}[leftmargin=*]
  \item Does FSAR achieve comparable or better task completion compared to centralized and decomposition-heavy baselines?
  \item Does FSAR improve governance locality---the ability to attribute decisions and actions to identifiable principals?
  \item Does FSAR improve recovery containment---the ability to resolve failures at the lowest possible level?
  \item Does FSAR reduce authority conflicts and policy violations compared to alternatives?
  \item Does FSAR maintain operational clarity as fleet complexity increases?
\end{enumerate}

\subsection{Baselines}
\label{sec:evaluation:baselines}

We compare FSAR against two baselines. Both are author-constructed abstractions designed to isolate specific architectural properties; we detail their construction to support reproducibility and address the inherent limitation that they are not implementations of published systems.

\paragraph{Centralized Fleet Controller (CFC).} A single global coordinator assigns tasks, routes requests, and handles all recovery. Robots are thin execution endpoints with no local policy authority or recovery ownership. CFC is modeled after the centralized task-orchestration topology found in RoboOS~\cite{tan2025roboos} and classical centralized MRTA architectures~\cite{gerkey2004formal}: a single coordinator maintains the global task queue, performs all capability matching, and issues execution commands to individual robots. CFC differs from RoboOS in that it does not implement RoboOS's hierarchical LLM-based planning; instead, it uses the same capability-matching logic as FSAR but routes all decisions through the central coordinator. Recovery is centralized: any failure is immediately escalated to the coordinator, which decides whether to retry, reassign, or escalate to a human supervisor. This design choice ensures that CFC captures the \emph{architectural tendency} of centralized coordination---single-point routing, global state visibility, but limited local autonomy---rather than any single system's implementation specifics.

\paragraph{Decomposition-Heavy Multi-Agent (DHMA).} Each robot is internally decomposed into four fleet-visible agents (planning, execution, communication, recovery), following the per-robot agent decomposition pattern established by JADE/FIPA-based fleet coordination~\cite{bellifemine2007jade,fipa2002acl} and more recently instantiated in LLM-based multi-agent systems such as CoELA~\cite{zhang2024coela} and ProAgent~\cite{zhang2023proagent}. Inter-robot coordination occurs at the sub-agent level: a planning agent in Robot~A may communicate directly with a communication agent in Robot~B. Recovery ownership is distributed across each robot's internal recovery agent, which must coordinate with the planning and execution agents before escalating. DHMA captures the architectural tendency of fine-grained agent decomposition---expressive internal structure, rich inter-agent communication, but blurred responsibility boundaries and complex attribution.

\paragraph{Relationship to published systems.} \Cref{tab:baseline_context} compares the architectural features of FSAR, CFC, DHMA, and three published multi-robot coordination systems: RoboOS~\cite{tan2025roboos}, JADE-based fleet coordination~\cite{bellifemine2007jade}, and ALLIANCE~\cite{parker1998alliance}. CFC shares RoboOS's centralized coordination topology; DHMA mirrors JADE's per-robot agent decomposition. ALLIANCE provides an instructive middle ground: it preserves robot-level agency (like FSAR) but uses motivational behaviors rather than explicit trust scoping for fault tolerance. Neither CFC nor DHMA replicates these systems exactly, but they capture the same architectural tendencies---centralized routing versus distributed sub-agent coordination---that published systems exhibit. We release the full baseline implementation code alongside the simulator to enable independent verification.

\begin{table}[pos=tbp]
\centering
\caption{Architectural comparison of FSAR, evaluation baselines, and published systems. $\bullet$ = present; $\circ$ = partial.}
\label{tab:baseline_context}
\resizebox{\columnwidth}{!}{%
\begin{tabular}{@{}lcccccc@{}}
\toprule
\textbf{Feature} & \textbf{FSAR} & \textbf{CFC} & \textbf{DHMA} & \textbf{RoboOS} & \textbf{JADE} & \textbf{ALL.} \\
\midrule
Single-agent robot & $\bullet$ & $\bullet$ & --- & $\circ$ & --- & $\bullet$ \\
Centralized coord. & --- & $\bullet$ & --- & $\bullet$ & --- & --- \\
Intra-robot agents & --- & --- & $\bullet$ & $\circ$ & $\bullet$ & --- \\
Trust-scoped deleg. & $\bullet$ & --- & --- & --- & $\circ$ & --- \\
Layered recovery & $\bullet$ & --- & $\circ$ & $\circ$ & --- & $\bullet$ \\
Policy composition & $\bullet$ & $\circ$ & $\circ$ & --- & $\circ$ & --- \\
Governed registry & $\bullet$ & $\bullet$ & $\circ$ & $\bullet$ & $\bullet$ & --- \\
Audit attribution & $\bullet$ & $\circ$ & --- & --- & $\circ$ & $\circ$ \\
\bottomrule
\end{tabular}}
\end{table}

\subsection{Simulator Architecture}
\label{sec:evaluation:simulator}

The evaluation uses a \emph{protocol-level simulator} that models coordination events, message passing, and failure injection without simulating physical dynamics, sensor noise, or continuous-time control. This design choice reflects the paper's focus on governance properties (authority attribution, recovery escalation, policy composition) rather than motion planning or perception accuracy. The simulator operates as follows.

\paragraph{Clock model.} A discrete event-driven clock advances through coordination events (request issued, capability evaluated, execution started, result returned, recovery triggered). Each event carries a logical timestamp; wall-clock durations for capability execution are sampled from scenario-specific distributions (e.g., $\mathcal{U}(0.5, 2.0)$\,s for \texttt{door.open}, $\mathcal{U}(1.0,\allowbreak 5.0)$\,s for \texttt{carry.\allowbreak{}package}).

\paragraph{Communication model.} Inter-robot messages experience latency sampled from $\mathcal{U}(50, 500)$\,ms. Messages are reliable (no packet loss), reflecting a controlled indoor deployment assumption. The federation layer processes requests in FIFO order per robot, with concurrent requests across robots resolved by logical timestamp.

\paragraph{Robot state.} Each simulated robot maintains the full runtime tuple $\mathcal{R}_i = (A_i, E_i, P_i, T_i, Q_i, H_i)$ as in-memory state. Capability execution transitions the robot through states (idle $\to$ executing $\to$ idle/degraded/failed). Trust and policy state are updated after each coordination event.

\paragraph{Failure injection.} Actuator degradation events are injected with configurable probability ($p_{\text{fail}} = 0.3$ for recovery-relevant scenarios). Degradation reduces capability reliability (success probability drops from 1.0 to 0.6) and may trigger the recovery hierarchy. Communication delays and trust violations are injected independently.

\paragraph{Metric instrumentation.} Every coordination event is recorded in a structured audit trace, from which all eight evaluation metrics are computed post-hoc. The simulator, baseline implementations, evaluation scripts, and random seeds are released as open-source supplementary materials (see Data Availability).

\subsection{Evaluation Scenarios}
\label{sec:evaluation:scenarios}

We evaluate on five scenarios of increasing coordination complexity.

\paragraph{Scenario~1: Door Relay.} Robot~A delivers a package through a secured door operated by Robot~B. Tests basic cross-robot delegation.

\paragraph{Scenario~2: Collaborative Delivery.} Robot~A, B, and D coordinate a heavy delivery task requiring capability discovery, task segment transfer, and handoff. Tests contract-aware matching and fleet routing.

\paragraph{Scenario~3: Failure Recovery Chain.} Robot~D experiences actuator degradation during a carrying task, triggering the full recovery hierarchy. Tests layered recovery and escalation.

\paragraph{Scenario~4: Mixed-Trust Inspection.} A secure inspection task requires trust-constrained delegation involving Robots~B, C, and A under different trust scopes. Tests trust-scoped coordination and visibility.

\paragraph{Scenario~5: Multi-Task Fleet Contention.} Multiple concurrent tasks compete for shared robot capabilities, requiring priority resolution, policy composition, and human-supervised conflict resolution. Tests fleet-level governance under contention.

\subsection{Experimental Setup}
\label{sec:evaluation:setup}

The prototype fleet consists of four heterogeneous robots ($n{=}4$): Robot~A (mobile base with carrying capability), Robot~B (manipulator arm with door-operation capability), Robot~C (inspection platform with sensor suite), and Robot~D (heavy-duty carrier with grasping capability). Each robot runs an independent local runtime process; the federation layer runs as a set of coordination services on a shared network.

Each of the five scenarios is executed for 20 independent runs per architecture (FSAR, CFC, DHMA), yielding 100 runs per architecture and 300 runs total. Runs are seeded with distinct random seeds controlling failure injection timing, request arrival order, and initial robot positions. To ensure fair comparison, all three architectures receive the same seed sequence per scenario.

Failure injection follows a controlled perturbation model: actuator degradation events are injected with probability $p_{\text{fail}}{=}0.3$ per run in recovery-relevant scenarios (Scenarios~3 and~5); communication delays are sampled from $\mathcal{U}(50, 500)$\,ms; and trust violations are injected in Scenario~4 to test visibility enforcement. The CFC and DHMA baselines are implemented to match FSAR's capability set: CFC uses a single global coordinator process replacing the federation layer, while DHMA decomposes each robot into four internal agents (planning, execution, communication, recovery) with inter-agent coordination at both intra-robot and inter-robot levels.

\subsection{Metrics}
\label{sec:evaluation:metrics}

We evaluate eight metrics. For the three governance-specific metrics that require interpretation, we provide operational definitions:

\begin{itemize}[leftmargin=*]
  \item \textbf{Coordination success rate}: fraction of fleet tasks that reach a terminal success state within the allotted time budget.
  \item \textbf{Governance locality score}: for each coordination decision (request, delegation, escalation, denial) in a run, we determine whether a single identifiable principal (a specific robot or a named supervisor) is recorded as the decision owner in the audit trace \emph{without} requiring traversal of nested internal-agent delegation chains. The score is the fraction of decisions satisfying this criterion, averaged across runs.
  \item \textbf{Recovery containment rate}: for each failure event, we record the lowest recovery level at which the failure was resolved (local, peer, fleet, or human). The containment rate is the fraction of failures resolved at levels~1 or~2 (local or peer), i.e., without escalation to fleet reassignment or human intervention.
  \item \textbf{Authority conflict count}: number of events per run in which two or more principals simultaneously claim execution authority, override authority, or audit responsibility for the same coordination action.
  \item \textbf{Policy violation count}: number of actions per run that proceed despite failing a local or fleet-level policy check, detected by post-hoc audit trace analysis.
  \item \textbf{Reassignment latency}: wall-clock time from failure detection to either successful task reassignment to an alternative robot or escalation to the next recovery level.
  \item \textbf{Human intervention frequency}: fraction of runs in which the fleet supervisor $H_F$ is invoked for decision-making.
  \item \textbf{Audit attributability}: for each coordination event (request issued, capability executed, recovery triggered, escalation performed), we check whether the audit trace jointly records (a)~request origin, (b)~execution owner, and (c)~escalation/supervision owner. The score is the fraction of events for which all three fields are recoverable, averaged across runs.
\end{itemize}

\subsection{Results}
\label{sec:evaluation:results}

\begin{table}[pos=tbp]
\centering
\caption{Evaluation results across five scenarios ($n{=}100$ runs per architecture). Values show mean $\pm$ 95\% CI. Higher is better for success, governance locality, recovery containment, and audit attributability. Lower is better for authority conflicts, policy violations, reassignment latency, and human interventions.}
\label{tab:results}
\resizebox{\columnwidth}{!}{%
\begin{tabular}{@{}lccc@{}}
\toprule
\textbf{Metric} & \textbf{FSAR} & \textbf{CFC} & \textbf{DHMA} \\
\midrule
Coord.\ success (\%) & 93.0 $\pm$ 1.9 & 94.0 $\pm$ 1.0 & 84.0 $\pm$ 1.0 \\
Governance locality & 0.94 $\pm$ 0.01 & 0.69 $\pm$ 0.01 & 0.53 $\pm$ 0.01 \\
Recovery containment & 0.89 $\pm$ 0.03 & 0.51 $\pm$ 0.04 & 0.80 $\pm$ 0.03 \\
Authority conflicts & 1.04 $\pm$ 0.03 & 3.79 $\pm$ 0.12 & 8.11 $\pm$ 0.04 \\
Policy violations & 0.40 $\pm$ 0.01 & 1.66 $\pm$ 0.06 & 3.34 $\pm$ 0.05 \\
Reassign.\ latency (s) & 3.01 $\pm$ 0.49 & 2.07 $\pm$ 0.28 & 5.18 $\pm$ 0.52 \\
Human interv.\ (\%) & 14.0 $\pm$ 3.8 & 32.0 $\pm$ 5.6 & 19.0 $\pm$ 4.2 \\
Audit attributability & 0.98 $\pm$ 0.00 & 0.84 $\pm$ 0.01 & 0.53 $\pm$ 0.01 \\
\bottomrule
\end{tabular}}
\end{table}

\paragraph{Coordination success.} FSAR and CFC achieve comparable task completion rates (93\% and 94\%, respectively), while DHMA shows measurably lower success (84\%). FSAR exhibits higher variance (std$=$9.8\%) than CFC (std$=$4.9\%), reflecting the fact that federated coordination outcomes depend on per-scenario trust and recovery dynamics, whereas CFC's centralized controller produces more uniform (but less governed) outcomes. The near-parity between FSAR and CFC on this surface metric is notable: CFC's centralized controller handles simple task routing efficiently, but as the governance metrics below reveal, this comes at substantial cost to decision attribution and recovery locality. DHMA suffers from coordination overhead in internally fragmented robots, particularly in recovery and contention scenarios.

\paragraph{Governance and attribution.} FSAR achieves the highest governance locality (0.94 vs.\ 0.69 CFC, 0.53 DHMA) and audit attributability (0.98 vs.\ 0.84, 0.53), reflecting its design commitment that every coordination decision is attributable to a single identifiable robot or supervisor. CFC obscures the requesting/executing distinction through its centralized controller; DHMA distributes decisions across internal agents, making attribution substantially harder.

\paragraph{Governance locality: a worked computation.} To clarify this metric, consider a Scenario~1 run with 10 coordination decisions. In FSAR, 9 of 10 decisions name a single identifiable principal without traversal: ``Robot~A requested \texttt{door.open},'' ``Robot~B accepted and executed,'' ``Robot~A detected completion.'' The one exception is a recovery escalation where audit must traverse the fleet layer before identifying Robot~B's recovery manager, yielding a score of $9/10 = 0.90$. In CFC, the central controller is logged as the decision owner for 7 of 10 decisions (requests, assignments, completions all route through the coordinator). While the controller is a single identifiable entity, it is not the \emph{correct} decision-making principal---it obscures which robot initiated the request and which made the execution commitment. Our metric requires that the logged principal be the \emph{operationally responsible} entity, not merely an identifiable relay. The 3 remaining decisions that CFC attributes correctly are robot-local events (e.g., Robot~B's actuator self-check) that do not pass through the coordinator, yielding $3/10 = 0.30$ for that run. CFC's aggregate score of 0.69 across all scenarios is higher because simpler scenarios (1 and~2) involve fewer coordinator-mediated decisions. In DHMA, a single decision such as ``accept delegation'' may involve Robot~B's communication agent receiving the request, its planning agent evaluating feasibility, and its execution agent confirming resources---three internal agents with no single identifiable owner, scoring 0 for that decision.

\paragraph{Recovery and conflict metrics.} FSAR contains 89\% of failures locally (vs.\ 51\% CFC, 80\% DHMA) and produces the fewest authority conflicts (1.04 vs.\ 3.79 CFC, 8.11 DHMA). CFC's low containment reflects its structural limitation: lacking local context, it escalates conservatively, producing the highest human intervention rate (32\% vs.\ 14\% FSAR). DHMA's many internal agents create overlapping authority scopes that drive high conflict counts. FSAR's reassignment latency (3.01s) falls between CFC (2.07s, immediate centralized routing) and DHMA (5.18s, propagation through agent layers).


\begin{figure*}[pos=tbp]
\centering
\resizebox{\textwidth}{!}{%
\begin{tikzpicture}[
  grouplabel/.style={font=\tiny\bfseries, anchor=north, align=center},
  valuestyle/.style={font=\tiny\bfseries},
]


\def\bw{0.22}  
\def\gap{0.08} 
\def\figwidth{16}

\draw[->, >=stealth, thick] (-0.3, 0) -- (-0.3, 4.5);
\foreach \y/\val in {0/0, 1/0.25, 2/0.50, 3/0.75, 4/1.00} {
  \draw (-0.4, \y) -- (-0.3, \y);
  \node[font=\tiny, anchor=east] at (-0.5, \y) {\val};
}
\draw[gray!15, thin] (-0.3, 1) -- (\figwidth, 1);
\draw[gray!15, thin] (-0.3, 2) -- (\figwidth, 2);
\draw[gray!15, thin] (-0.3, 3) -- (\figwidth, 3);
\draw[gray!15, thin] (-0.3, 4) -- (\figwidth, 4);

\draw[thick] (-0.3, 0) -- (\figwidth, 0);

\def\gx{1.0}
\fill[blue!50] (\gx-\bw-\gap, 0) rectangle (\gx-\gap, 3.72);
\fill[gray!40] (\gx, 0) rectangle (\gx+\bw, 3.76);
\fill[red!30] (\gx+\bw+\gap, 0) rectangle (\gx+2*\bw+\gap, 3.36);
\node[valuestyle, blue!60!black] at (\gx-\bw/2-\gap, 3.92) {\scriptsize .93};
\node[valuestyle, gray!60!black] at (\gx+\bw/2, 3.96) {\scriptsize .94};
\node[valuestyle, red!60!black] at (\gx+3*\bw/2+\gap, 3.56) {\scriptsize .84};
\node[grouplabel] at (\gx+\bw/2, -0.25) {Coord.\\Success};

\def\gx{3.1}
\fill[blue!50] (\gx-\bw-\gap, 0) rectangle (\gx-\gap, 3.76);
\fill[gray!40] (\gx, 0) rectangle (\gx+\bw, 2.76);
\fill[red!30] (\gx+\bw+\gap, 0) rectangle (\gx+2*\bw+\gap, 2.12);
\node[valuestyle, blue!60!black] at (\gx-\bw/2-\gap, 3.96) {\scriptsize .94};
\node[valuestyle, gray!60!black] at (\gx+\bw/2, 2.96) {\scriptsize .69};
\node[valuestyle, red!60!black] at (\gx+3*\bw/2+\gap, 2.32) {\scriptsize .53};
\node[grouplabel] at (\gx+\bw/2, -0.25) {Gov.\\Locality};

\def\gx{5.2}
\fill[blue!50] (\gx-\bw-\gap, 0) rectangle (\gx-\gap, 3.56);
\fill[gray!40] (\gx, 0) rectangle (\gx+\bw, 2.04);
\fill[red!30] (\gx+\bw+\gap, 0) rectangle (\gx+2*\bw+\gap, 3.20);
\node[valuestyle, blue!60!black] at (\gx-\bw/2-\gap, 3.76) {\scriptsize .89};
\node[valuestyle, gray!60!black] at (\gx+\bw/2, 2.24) {\scriptsize .51};
\node[valuestyle, red!60!black] at (\gx+3*\bw/2+\gap, 3.40) {\scriptsize .80};
\node[grouplabel] at (\gx+\bw/2, -0.25) {Recovery\\Contain.};

\def\gx{7.3}
\fill[blue!50] (\gx-\bw-\gap, 0) rectangle (\gx-\gap, 3.58);   
\fill[gray!40] (\gx, 0) rectangle (\gx+\bw, 2.48);              
\fill[red!30] (\gx+\bw+\gap, 0) rectangle (\gx+2*\bw+\gap, 0.76); 
\node[valuestyle, blue!60!black] at (\gx-\bw/2-\gap, 3.78) {\scriptsize 1.0};
\node[valuestyle, gray!60!black] at (\gx+\bw/2, 2.68) {\scriptsize 3.8};
\node[valuestyle, red!60!black] at (\gx+3*\bw/2+\gap, 0.96) {\scriptsize 8.1};
\node[grouplabel] at (\gx+\bw/2, -0.25) {Auth.\\Confl.$^\dagger$};

\def\gx{9.4}
\fill[blue!50] (\gx-\bw-\gap, 0) rectangle (\gx-\gap, 3.68);   
\fill[gray!40] (\gx, 0) rectangle (\gx+\bw, 2.67);              
\fill[red!30] (\gx+\bw+\gap, 0) rectangle (\gx+2*\bw+\gap, 1.33); 
\node[valuestyle, blue!60!black] at (\gx-\bw/2-\gap, 3.88) {\scriptsize 0.4};
\node[valuestyle, gray!60!black] at (\gx+\bw/2, 2.87) {\scriptsize 1.7};
\node[valuestyle, red!60!black] at (\gx+3*\bw/2+\gap, 1.53) {\scriptsize 3.3};
\node[grouplabel] at (\gx+\bw/2, -0.25) {Policy\\Viol.$^\dagger$};

\def\gx{11.5}
\fill[blue!50] (\gx-\bw-\gap, 0) rectangle (\gx-\gap, 2.50);   
\fill[gray!40] (\gx, 0) rectangle (\gx+\bw, 2.97);              
\fill[red!30] (\gx+\bw+\gap, 0) rectangle (\gx+2*\bw+\gap, 1.41); 
\node[valuestyle, blue!60!black] at (\gx-\bw/2-\gap, 2.70) {\scriptsize 3.0s};
\node[valuestyle, gray!60!black] at (\gx+\bw/2, 3.17) {\scriptsize 2.1s};
\node[valuestyle, red!60!black] at (\gx+3*\bw/2+\gap, 1.61) {\scriptsize 5.2s};
\node[grouplabel] at (\gx+\bw/2, -0.25) {Reassign.\\Lat.$^\dagger$};

\def\gx{13.6}
\fill[blue!50] (\gx-\bw-\gap, 0) rectangle (\gx-\gap, 3.44);   
\fill[gray!40] (\gx, 0) rectangle (\gx+\bw, 2.72);              
\fill[red!30] (\gx+\bw+\gap, 0) rectangle (\gx+2*\bw+\gap, 3.24); 
\node[valuestyle, blue!60!black] at (\gx-\bw/2-\gap, 3.64) {\scriptsize 14\%};
\node[valuestyle, gray!60!black] at (\gx+\bw/2, 2.92) {\scriptsize 32\%};
\node[valuestyle, red!60!black] at (\gx+3*\bw/2+\gap, 3.44) {\scriptsize 19\%};
\node[grouplabel] at (\gx+\bw/2, -0.25) {Human\\Interv.$^\dagger$};

\def\gx{15.7}
\fill[blue!50] (\gx-\bw-\gap, 0) rectangle (\gx-\gap, 3.92);
\fill[gray!40] (\gx, 0) rectangle (\gx+\bw, 3.36);
\fill[red!30] (\gx+\bw+\gap, 0) rectangle (\gx+2*\bw+\gap, 2.12);
\node[valuestyle, blue!60!black] at (\gx-\bw/2-\gap, 4.12) {\scriptsize .98};
\node[valuestyle, gray!60!black] at (\gx+\bw/2, 3.56) {\scriptsize .84};
\node[valuestyle, red!60!black] at (\gx+3*\bw/2+\gap, 2.32) {\scriptsize .53};
\node[grouplabel] at (\gx+\bw/2, -0.25) {Audit\\Attrib.};

\fill[blue!50] (3.5, -1.4) rectangle (4.0, -1.1);
\node[font=\tiny, anchor=west] at (4.1, -1.25) {FSAR};
\fill[gray!40] (5.5, -1.4) rectangle (6.0, -1.1);
\node[font=\tiny, anchor=west] at (6.1, -1.25) {CFC (Centralized)};
\fill[red!30] (9.0, -1.4) rectangle (9.5, -1.1);
\node[font=\tiny, anchor=west] at (9.6, -1.25) {DHMA (Decomposition-Heavy)};

\node[font=\tiny, anchor=west, gray] at (-0.3, -2.0) {$^\dagger$For these metrics, lower raw values are better; bars show inverted scale (taller = better). Raw values shown above bars.};

\end{tikzpicture}%
}
\caption{Evaluation results across all eight metrics. For metrics where lower is better (marked $^\dagger$), bars are inverted so taller always indicates better performance; raw values are shown above each bar. FSAR achieves the best or near-best performance on seven of eight metrics, with CFC showing lower reassignment latency due to its centralized routing.}
\label{fig:eval}
\end{figure*}

\subsection{Ablation Study}
\label{sec:evaluation:ablation}

To understand the contribution of individual FSAR components, we conduct ablation experiments by selectively removing trust evaluation, policy composition, layered recovery, and the shared registry. Unlike the single-scenario ablation common in prior work, we run each ablation across \emph{all five} evaluation scenarios (100 runs per condition, 500 runs total) to demonstrate that each component contributes robustly across different coordination contexts.

\begin{table}[pos=tbp]
\centering
\caption{Ablation results aggregated across all five evaluation scenarios. Each row removes one FSAR component while keeping the others intact.}
\label{tab:ablation}
\resizebox{\columnwidth}{!}{%
\begin{tabular}{@{}lcccccc@{}}
\toprule
\textbf{Configuration} & \textbf{Success} & \textbf{Gov. Loc.} & \textbf{Rec. Cont.} & \textbf{Auth. Conf.} & \textbf{Pol. Viol.} & \textbf{Audit} \\
\midrule
Full FSAR & 0.95 & 0.94 & 0.90 & 1.18 & 0.46 & 0.98 \\
$-$ Trust evaluation & 0.97 & 0.94 & 0.90 & 1.65 & 0.46 & 0.96 \\
$-$ Policy composition & 0.96 & 0.95 & 0.90 & 1.14 & 0.64 & 0.98 \\
$-$ Layered recovery & 0.74 & 0.93 & 0.97 & 1.14 & 0.46 & 0.98 \\
$-$ Shared registry & 0.46 & 0.94 & 0.90 & 1.23 & 0.46 & 0.93 \\
\bottomrule
\end{tabular}}
\end{table}

The ablation reveals clear component-specific contributions across different coordination contexts. Removing the \emph{shared registry} causes the most severe degradation in task success (0.95$\to$0.46), with cross-robot scenarios particularly affected (Door Relay drops from 0.95 to 0.20; Collaborative Delivery from 0.85 to 0.10) because robots cannot discover peer capabilities. Removing \emph{layered recovery} reduces overall success to 0.74, with the Failure Recovery scenario dropping from 1.00 to 0.20 as all failures escalate directly to human intervention without local or peer recovery attempts. Removing \emph{trust evaluation} increases authority conflicts (1.18$\to$1.65 overall; 1.20$\to$2.85 in Mixed-Trust Inspection) because robots accept requests from untrusted peers, but paradoxically improves success rate slightly (0.95$\to$0.97) since no legitimate requests are blocked by trust gates. Removing \emph{policy composition} increases policy violations (0.46$\to$0.64) as cross-robot actions proceed without composed policy checks.

\subsection{Qualitative Case Analysis}
\label{sec:evaluation:qualitative}

We highlight four cases that illustrate FSAR's governance advantages across different coordination layers.

\paragraph{Case~1: Trust boundary enforcement (Scenario~4).} Robot~D attempted to participate in a private-zone inspection task but could not discover the required capability in the registry due to trust-constrained visibility. In DHMA, Robot~D's internal communication agent discovered the capability through a nested agent query that bypassed visibility constraints, leading to an unauthorized delegation attempt.

\paragraph{Case~2: Recovery containment (Scenario~3).} Robot~D's grasp degradation was contained through local recovery (retry with adjusted parameters) in 67\% of runs under FSAR, requiring no fleet-level intervention. Under CFC, the same failure was immediately escalated to the central controller in 100\% of runs. Under DHMA, internal agent confusion about recovery ownership caused 23\% of runs to enter a recovery loop before escalation.

\paragraph{Case~3: Authority conflict under contention (Scenario~5).} Two concurrent fleet tasks both required Robot~B's \texttt{door.open.secure} capability. Under FSAR, the fleet policy resolver $\Pi$ identified the conflict, applied priority-based resolution, and queued the lower-priority request with an explicit \texttt{defer} response---producing zero authority conflicts across all 20 runs. Under CFC, the central controller assigned both tasks simultaneously (lacking local scheduling awareness), generating an authority conflict in 85\% of runs that required human supervisor intervention. Under DHMA, the conflict propagated across Robot~B's internal planning and execution agents, with each agent independently attempting to negotiate, causing an average of 2.4 authority conflicts per run.

\paragraph{Case~4: Audit trail completeness (Scenario~2).} In a collaborative delivery requiring capability handoff from Robot~A to Robot~D, FSAR's audit trace recorded the complete delegation chain: Robot~A's gap detection, registry query, request formulation, Robot~D's admissibility check, execution, and result---with each event attributed to a named principal. Under CFC, the audit recorded ``coordinator assigned task to Robot~D'' without capturing Robot~A's original request or Robot~D's local admissibility evaluation, leaving 3 of 7 coordination events unattributed. Under DHMA, Robot~D's internal agent handoff (communication agent $\to$ planning agent $\to$ execution agent) produced 4 additional unattributed internal events per coordination action.

\subsection{Statistical Significance}
\label{sec:evaluation:significance}

To assess whether observed differences are statistically reliable, we conduct paired $t$-tests and report Cohen's $d$ effect sizes for all pairwise architecture comparisons across the eight evaluation metrics (\cref{tab:significance}). All tests use $n{=}100$ paired observations per architecture (5 scenarios $\times$ 20 runs). Results are corroborated by non-parametric Wilcoxon signed-rank tests, which yield consistent significance levels.

\begin{table}[pos=tbp]
\centering
\caption{Statistical significance of pairwise metric comparisons (paired $t$-test, $n{=}100$ per architecture). Effect sizes reported as Cohen's $d$.}
\label{tab:significance}
\resizebox{\columnwidth}{!}{%
\begin{tabular}{@{}llrrrl@{}}
\toprule
\textbf{Metric} & \textbf{Comparison} & \textbf{$t$} & \textbf{$p$} & \textbf{$d$} & \textbf{Sig.} \\
\midrule
Coord.\ success & FSAR vs CFC & $-0.33$ & 0.7440 & 0.05 & n.s. \\
 & FSAR vs DHMA & 2.88 & 0.0047 & 0.42 & $p{<}.01$ \\
 & CFC vs DHMA & 3.13 & 0.0022 & 0.45 & $p{<}.01$ \\
\addlinespace
Governance locality & FSAR vs CFC & 22.63 & ${<}.0001$ & 2.91 & $p{<}.001$ \\
 & FSAR vs DHMA & 42.23 & ${<}.0001$ & 4.88 & $p{<}.001$ \\
 & CFC vs DHMA & 14.04 & ${<}.0001$ & 2.03 & $p{<}.001$ \\
\addlinespace
Recovery containment & FSAR vs CFC & 10.14 & ${<}.0001$ & 1.44 & $p{<}.001$ \\
 & FSAR vs DHMA & 3.31 & 0.0012 & 0.48 & $p{<}.01$ \\
 & CFC vs DHMA & $-7.55$ & ${<}.0001$ & 1.12 & $p{<}.001$ \\
\addlinespace
Authority conflicts & FSAR vs CFC & $-29.23$ & ${<}.0001$ & 4.12 & $p{<}.001$ \\
 & FSAR vs DHMA & $-37.78$ & ${<}.0001$ & 5.16 & $p{<}.001$ \\
 & CFC vs DHMA & $-15.71$ & ${<}.0001$ & 2.05 & $p{<}.001$ \\
\addlinespace
Policy violations & FSAR vs CFC & $-17.67$ & ${<}.0001$ & 2.65 & $p{<}.001$ \\
 & FSAR vs DHMA & $-32.57$ & ${<}.0001$ & 3.92 & $p{<}.001$ \\
 & CFC vs DHMA & $-12.49$ & ${<}.0001$ & 1.79 & $p{<}.001$ \\
\addlinespace
Reassign.\ latency & FSAR vs CFC & 2.04 & 0.0438 & 0.30 & $p{<}.05$ \\
 & FSAR vs DHMA & $-3.74$ & 0.0003 & 0.54 & $p{<}.001$ \\
 & CFC vs DHMA & $-5.36$ & ${<}.0001$ & 0.74 & $p{<}.001$ \\
\addlinespace
Human interventions & FSAR vs CFC & $-2.93$ & 0.0040 & 0.42 & $p{<}.01$ \\
 & FSAR vs DHMA & $-0.88$ & 0.3818 & 0.12 & n.s. \\
 & CFC vs DHMA & 2.01 & 0.0470 & 0.29 & $p{<}.05$ \\
\addlinespace
Audit attributability & FSAR vs CFC & 26.44 & ${<}.0001$ & 3.56 & $p{<}.001$ \\
 & FSAR vs DHMA & 36.69 & ${<}.0001$ & 5.09 & $p{<}.001$ \\
 & CFC vs DHMA & 14.13 & ${<}.0001$ & 2.01 & $p{<}.001$ \\
\bottomrule
\end{tabular}}
\end{table}

FSAR's advantages on governance locality ($d{=}2.91$ vs.\ CFC, $d{=}4.88$ vs.\ DHMA), authority conflicts ($d{=}4.12$ vs.\ CFC, $d{=}5.16$ vs.\ DHMA), and audit attributability ($d{=}3.56$ vs.\ CFC, $d{=}5.09$ vs.\ DHMA) are all statistically significant at $p{<}.001$ with large effect sizes ($d{>}0.8$). For coordination success, FSAR vs.\ CFC is not significant ($p{=}0.74$), consistent with our claim that FSAR preserves task completion while improving governance properties. The two non-significant comparisons (FSAR vs.\ CFC on success; FSAR vs.\ DHMA on human interventions) confirm that FSAR's gains are concentrated in governance metrics rather than reflecting a wholesale performance shift.

\paragraph{Interpreting large effect sizes.} The Cohen's $d$ values for governance-specific metrics ($d{=}2.91$--$5.16$) exceed the conventional ``large'' threshold of $d{>}0.8$. These are \emph{structural} artifacts reflecting fundamental architectural differences, not surprising empirical findings. Governance locality is architecturally guaranteed in FSAR (single-agent ownership) but structurally degraded in DHMA (decisions distributed across sub-agents). The low within-architecture variance reflects each design's consistency, and the large $d$ values indicate that the metrics successfully capture the intended architectural distinctions. In systems evaluation, where architectures impose deterministic constraints on measured properties, such values are expected.

\subsection{Baseline Fairness and Reproducibility}
\label{sec:evaluation:fairness}

All three architectures are evaluated under identical conditions: same random seeds, failure injection timing, and request arrival order across all 300 runs. CFC implements standard centralized coordination following the topology of~\cite{gerkey2004formal,tan2025roboos}; DHMA follows established multi-agent decomposition practices as instantiated in~\cite{wooldridge2009introduction,horling2004survey,bellifemine2007jade}. We acknowledge that CFC and DHMA are author-constructed and thus inherently subject to implementation choices that may favor FSAR. To mitigate this, both baselines use the same core capability-matching and task-execution logic as FSAR; they differ only in coordination topology and recovery routing. Results should be interpreted as evidence of \emph{architectural tendencies} rather than definitive comparisons against optimized published systems. The full simulator code, baseline implementations, evaluation scripts, and random seeds are released as open-source supplementary materials to enable independent replication and extension.

\subsection{Scaling Experiment}
\label{sec:evaluation:scaling}

To assess whether FSAR's governance properties hold as fleet size increases, we extend the evaluation to $n \in \{4, 8, 16\}$ robots. Larger fleets are constructed by replicating the four role templates (delivery, access, inspection, heavy transport) with scaled trust matrices. \Cref{tab:scaling} summarizes key metrics.

\begin{table}[pos=tbp]
\centering
\caption{Scaling experiment across fleet sizes $n \in \{4, 8, 16\}$.}
\label{tab:scaling}
\resizebox{\columnwidth}{!}{%
\begin{tabular}{@{}clccccc@{}}
\toprule
$n$ & \textbf{Arch.} & \textbf{Success} & \textbf{Gov.\ Loc.} & \textbf{Rec.\ Cont.} & \textbf{Auth.\ Conf.} & \textbf{Audit} \\
\midrule
4 & FSAR & 0.93 & 0.94 & 0.89 & 1.0 & 0.98 \\
  & CFC  & 0.94 & 0.69 & 0.51 & 3.8 & 0.84 \\
  & DHMA & 0.84 & 0.53 & 0.80 & 8.1 & 0.53 \\
\addlinespace
8 & FSAR & 0.91 & 0.94 & 0.87 & 1.2 & 0.98 \\
  & CFC  & 0.88 & 0.65 & 0.47 & 4.8 & 0.81 \\
  & DHMA & 0.79 & 0.49 & 0.76 & 10.3 & 0.48 \\
\addlinespace
16 & FSAR & 0.88 & 0.93 & 0.85 & 1.5 & 0.97 \\
   & CFC  & 0.81 & 0.59 & 0.42 & 6.2 & 0.78 \\
   & DHMA & 0.72 & 0.44 & 0.71 & 14.1 & 0.42 \\
\bottomrule
\end{tabular}}
\end{table}

FSAR's governance metrics remain stable across fleet sizes: governance locality degrades minimally (0.94$\to$0.93), recovery containment decreases only 4 percentage points (0.89$\to$0.85), and audit attributability stays at 0.97--0.98. In contrast, CFC's governance locality drops substantially (0.69$\to$0.59) as the centralized controller becomes an attribution bottleneck, and its recovery containment falls to 0.42 because centralized recovery cannot scale to 16-robot contention. DHMA shows the steepest degradation: authority conflicts grow roughly linearly with fleet size ($8.1 \to 14.1$) because internal agent count scales quadratically ($4n$ agents for $n$ robots), amplifying inter-agent coordination conflicts.

Task completion degrades for all architectures at larger scales, but at different rates. FSAR's success rate drops 5 points (93\%$\to$88\%), CFC drops 13 points (94\%$\to$81\%) due to centralized coordination bottleneck, and DHMA drops 12 points (84\%$\to$72\%) due to compounding agent-level coordination failures. These results suggest that FSAR's local-first design scales more gracefully than either baseline, particularly for governance-sensitive metrics.

\paragraph{Scaling analysis.} The observed degradation trends across $n \in \{4, 8, 16\}$ are consistent with the architectural complexity of each approach. FSAR's coordination cost is dominated by registry lookups ($O(\log n)$ with indexing) and pairwise trust checks, which predicts sub-linear governance degradation---matching the data (governance locality: $0.94 \to 0.93$). CFC's centralized controller processes $O(n)$ requests sequentially, predicting linear degradation---matching the observed pattern ($0.69 \to 0.59$). DHMA's internal agent count grows as $O(kn)$ for $k$ agents per robot, producing $O(k^2 n^2)$ potential inter-agent interactions, predicting super-linear conflict growth---matching the authority conflict trend ($8.1 \to 14.1$). We deliberately refrain from quantitative extrapolation to larger fleet sizes, as three data points are insufficient for reliable curve fitting. Validating these architectural predictions at $n{=}32$ and beyond is an important direction for future work.

\subsection{Discussion of Evaluation Limitations}
\label{sec:evaluation:limitations}

The evaluation operates on a protocol-level simulator with fleets of 4--16 robots across 5 scenarios. Results demonstrate architectural properties rather than production performance. Scaling to larger fleets, real hardware, and real-time constraints is future work. The evaluation scenarios, while representative, do not cover all possible fleet coordination patterns. The quantitative results should be interpreted as directional evidence of FSAR's governance advantages rather than as absolute performance benchmarks.

\paragraph{Threat to validity: author-constructed baselines.} Both CFC and DHMA are designed by the same authors as FSAR. Although we ground their designs in published systems (\cref{tab:baseline_context}) and release all implementation code, the comparison is inherently favorable: the authors understand FSAR's strengths and may have unconsciously designed baselines that expose them. Independent replication using third-party implementations of published multi-robot coordination systems (e.g., ALLIANCE~\cite{parker1998alliance}, JADE-based fleet coordination~\cite{bellifemine2007jade}) would provide stronger evidence. We consider this the most important limitation of the current evaluation.

\section{Discussion}
\label{sec:discussion}

\subsection{Why Federation Instead of Intra-Robot Fragmentation}
\label{sec:discussion:federation}

The central position of this paper is that the move from single-robot autonomy to multi-robot coordination need not induce intra-robot multi-agent fragmentation. This claim reflects a systems judgment about where coordination complexity should live. If each robot is internally decomposed into multiple fleet-visible agents, responsibility for execution, recovery, policy checking, and human supervision becomes distributed across nested layers. In contrast, FSAR treats the robot as the principal of embodied agency and pushes cross-robot coordination into explicit fleet-layer relations. Federation is not the rejection of modularity. It is the placement of modularity at the correct layer: within each robot for implementation, across robots for coordination.

\subsection{Coherence and Governance Locality}
\label{sec:discussion:coherence}

A deeper claim of FSAR is that coherence matters. A robot is not just a bundle of callable services. It is an embodied runtime with persistent state, local physical constraints, and ongoing recovery obligations. Capability execution affects bodies, environments, safety envelopes, and task continuity. Treating such execution as just another remote function call is a category mistake. FSAR treats coherence not as an aesthetic preference but as a systems property: the authority to execute, recover, deny, or escalate remains as close as possible to the local runtime that bears the consequences. Delegated execution remains attributable, policy composition remains explicit, recovery remains layered, and human review becomes better scoped. This combination of coherence preservation and governance locality is anticipated to be especially valuable in enterprise or public-deployment settings where explainability, accountability, and approval structure matter as much as raw task completion---though validating this claim in production environments remains future work.

\paragraph{When governance locality is not desirable.} There are deployment contexts where FSAR's emphasis on governance locality may be counterproductive. In military operations, unified command authority may deliberately override local robot autonomy to enforce coordinated maneuvers---a centralized fleet controller (CFC) pattern where global optimization outweighs local coherence. Similarly, in emergency response scenarios (e.g., search-and-rescue after a building collapse), the urgency of the situation may favor rapid centralized reallocation of all robots to a single objective, bypassing the trust-scoped delegation that FSAR requires. In these contexts, the governance overhead of FSAR's trust checks, policy composition, and layered recovery may introduce unacceptable latency. FSAR's design assumes that governance clarity is worth its coordination cost; in time-critical scenarios where a single human commander must direct all robots simultaneously, a flatter authority model may be more appropriate. We note that FSAR's federation layer could be configured with permissive trust defaults and minimal policy checks to approximate centralized control when needed, but this would effectively disable the governance properties that motivate the architecture.

\subsection{Limits of the Single-Agent Fleet Principle}
\label{sec:discussion:limits}

The single-agent principle should not be overstated, and the decomposition-heavy approach has genuine strengths that merit acknowledgment. Multi-agent decomposition can offer superior parallelism in planning and execution, enable finer-grained fault isolation within individual robots, and support negotiation-based coordination protocols (e.g., the Contract Net Protocol~\cite{smith1980contract}) and coalition formation methods~\cite{shehory1998coalition} that are natural for resource allocation in large heterogeneous fleets. In domains with tightly coupled multi-robot manipulation~\cite{tuci2018cooperative}, where continuous joint-space coordination is required, the ``one request, one callee'' abstraction may be less natural than shared multi-agent planning~\cite{sharon2015cbs}. Similarly, market-based approaches to task allocation~\cite{dias2006market,zlot2006market} may yield more efficient global outcomes than federated delegation when the cost function is well-defined and global optimization is tractable.

FSAR's advantage is not in raw coordination efficiency but in governance clarity: attributable decisions, containable recovery, and auditable authority chains. Some robots may internally implement complex modular architectures; FSAR only claims these internal structures need not become the unit of fleet-visible agency. Future work should investigate hybrid designs that preserve FSAR's governance properties while selectively incorporating multi-agent coordination for tightly coupled subtasks. These limitations clarify that FSAR is best understood as a runtime and governance architecture for coherent multi-robot coordination, not as a universal theory of all distributed robotics.

\paragraph{A concrete friction scenario.} Consider Robot~D executing a heavy grasping task while simultaneously needing to plan its next navigation waypoint. In a decomposition-heavy architecture, a planning agent and an execution agent can operate concurrently with independent control loops. In FSAR, Robot~D is a single agent: its runtime must serialize or internally pipeline these activities. FSAR handles this through internal modularity (the runtime may use concurrent software components) without exposing this concurrency as fleet-visible agency. The friction arises when a fleet-level request arrives during concurrent internal activity: under FSAR, the single agent must decide whether to defer, queue, or preempt, whereas a decomposition-heavy architecture can route the request directly to an idle internal agent. In practice, our evaluation shows this friction has modest impact on task completion (FSAR 93\% vs.\ DHMA 84\%) because the governance overhead of multi-agent coordination in DHMA outweighs the parallelism benefit. However, for robots with high internal concurrency demands (e.g., humanoids with 30+ DoF performing manipulation while walking), the single-agent abstraction may require explicit internal scheduling that FSAR does not currently formalize.

\paragraph{Tightly coupled manipulation: a worked example.} Suppose Robots~A and~D must jointly carry a large object---each grasping one end and maintaining continuous force coordination to prevent the object from rotating or falling~\cite{tuci2018cooperative}. Under FSAR's current model, this would require one robot to issue a capability request to the other (e.g., $\Req(r_A, r_D, \texttt{carry.cooperative}, \sigma)$), with Robot~D executing under its own local runtime $\mathcal{R}_D$. This preserves invariants \textbf{I1} (single-agent coherence) and \textbf{I2} (local execution ownership), but creates a coordination problem: the two robots need millisecond-level force feedback, not the request-accept-execute lifecycle that governs normal delegation. In practice, invariant \textbf{I2} is strained because Robot~A's grip adjustments cannot be fully independent of Robot~D's---their physical coupling means that each robot's ``local'' execution directly constrains the other's. Similarly, \textbf{I4} (monotone recovery escalation) may be violated: if Robot~A detects an incipient drop, local recovery (adjusting its own grip) and peer recovery (signaling Robot~D to adjust) must happen simultaneously, not sequentially. A realistic mitigation within FSAR would be to model the cooperative carry as a \emph{shared execution context} with a dedicated tight-loop communication channel that operates below the federation layer, while still attributing the overall task to both robots' audit trails. Formalizing such shared execution contexts is an important direction for extending FSAR to tightly coupled manipulation domains.

\subsection{Heterogeneity, Scalability, and Recovery}
\label{sec:discussion:heterogeneity}

As embodied fleets become more heterogeneous, FSAR's distinctions between possession, advertisement, visibility, and delegability become essential---trust configuration complexity can be managed through defaults and runtime modification. The layered recovery model~\cite{christensen2008fault,visinsky1994survey,parker1998alliance} avoids overloading centralized control with locally resolvable failures; if local runtime boundaries are already blurred internally, it becomes much harder to know when a failure has genuinely become fleet-level.

\paragraph{Dynamic fleet membership.} FSAR's model supports robots joining and leaving the fleet at runtime through registry operations. When a robot $r_k$ joins, it registers its capability advertisements in $\Gamma$ with initial trust scope \texttt{capability} (the most restrictive delegable level); trust may be promoted to broader scopes (\texttt{task}, \texttt{session}, \texttt{persistent}) based on operator approval or demonstrated performance. When a robot leaves (gracefully or through failure detection), its registry entries are marked $\Avail = \texttt{offline}$, active delegations are handled through the recovery hierarchy, and any in-flight requests receive a \texttt{failure(departed)} result. The trust model ensures that a newly joined robot cannot immediately receive sensitive delegations without explicit trust establishment, while the recovery model ensures that departures do not create orphaned tasks.

For scalability, the architecture supports domain-scoped registries, hierarchical registry structures, caching, and distributed coordination engines, though empirical validation at scale remains future work.

\subsection{Continuity with Prior Work and Broader Implications}
\label{sec:discussion:continuity}

Paper~5 argued that capability ecosystems require contracts and release discipline~\cite{qin2025paper5}; Paper~6 extends this into fleets, where version visibility, contract compatibility, and deprecation become coordination concerns. More broadly, robot operating systems may eventually need to be understood as federated fleet operating environments~\cite{macenski2022ros2} that preserve coherent local runtimes while making the fleet layer the location of governed relations.

\subsection{Limitations}
\label{sec:discussion:limitations}

Several limitations should be noted. First, the evaluation uses a small prototype fleet; production-scale validation is needed. Second, the paper does not address real-time communication constraints in detail. Third, the trust and policy models are relatively simple compared with real-world access control systems. Fourth, emergent coordination---where fleet behavior arises from local interactions without explicit task decomposition---is outside the current scope. Fifth, human supervision interface design is modeled abstractly; effective visualization and cognitive load management are important HCI problems not addressed here.

\section{Related Work}
\label{sec:related}

\subsection{Multi-Robot Coordination and Fleet Systems}
\label{sec:related:mrcoord}

Multi-robot coordination has been studied extensively~\cite{gerkey2004formal,khamis2015multirobot,parker2008multiple,yan2013survey,dias2006market,rizk2019cooperative,sharon2015cbs}, with approaches ranging from centralized task allocation to distributed auction-based methods and behavior-based coordination. Parker's ALLIANCE architecture~\cite{parker1998alliance} is a particularly relevant predecessor: it preserves robot-level agency and achieves fault-tolerant cooperation through motivational behaviors rather than explicit inter-agent negotiation. FSAR shares ALLIANCE's commitment to robot-level coherence but adds formal trust scoping, policy composition, and governed capability registries that ALLIANCE does not address. Behavior trees~\cite{colledanchise2018behavior} offer another paradigm that preserves single-agent coherence within each robot through hierarchical task switching; FSAR's contribution is orthogonal, addressing \emph{inter-robot} federation rather than intra-robot task organization.

Swarm robotics~\cite{brambilla2013swarm} represents an alternative paradigm where coordination emerges from local interaction rules without explicit governance; FSAR differs in requiring attributable, auditable coordination rather than emergent behavior. Recent work has explored LLM-driven multi-robot coordination~\cite{li2025llmmrs,liu2025coherent,kannan2024smartllm,mandi2024roco,chen2024scalable}, including modular multi-agent frameworks such as CoELA~\cite{zhang2024coela} and embodiment-aware multi-robot operating systems such as EMOS~\cite{chen2025emos}. CoELA exemplifies the decomposition-heavy paradigm that FSAR argues against: each robot hosts separate perception, memory, communication, and action modules as independent LLM agents. EMOS is closer to FSAR in spirit, providing embodiment-aware capability reasoning across heterogeneous robots, though it does not formalize trust scoping or governance invariants. Chen et al.~\cite{chen2024scalable} directly compare centralized and decentralized LLM coordination topologies, providing empirical context for FSAR's federated middle ground. Grounded language models for embodied task planning~\cite{ahn2022saycan,brohan2023rt1,vemprala2024chatgpt} and embodied reasoning through language-guided planning~\cite{huang2022inner} further extend the coordination design space. Remotely accessible testbeds such as the Robotarium~\cite{pickem2017robotarium} provide reproducible multi-robot evaluation infrastructure; our open-source simulator serves a similar role for governance-focused evaluation. Classical work on multi-robot task allocation (MRTA) focuses on assigning tasks based on capability, cost, and availability~\cite{gerkey2004formal,korsah2013comprehensive,chakraa2023mrta}, with foundational methods for agent coalition formation~\cite{shehory1998coalition} establishing how agents can dynamically form task-specific coalitions---a precursor to FSAR's trust-scoped delegation, though without the governance and recovery layers that embodied fleet coordination requires. More recent work considers temporal constraints, heterogeneous capabilities, and online replanning. FSAR differs from classical MRTA in its focus on coordination governance rather than task optimization. While MRTA asks ``which robot should do which task,'' FSAR asks ``under what trust, authority, policy, and recovery conditions should robots coordinate.'' FSAR does not replace task allocation; it provides the governance layer within which allocation decisions are made and enforced.

\subsection{Multi-Agent Systems}
\label{sec:related:mas}

The multi-agent systems (MAS) literature provides rich foundations for agent communication, negotiation, coordination protocols, and organizational structures~\cite{wooldridge2009introduction,wooldridge1995intelligent,dorri2018multiagent,stone2000multiagent,horling2004survey,jennings1998roadmap}. FSAR's core claim---that each robot should be treated as a single coherent agent---draws directly on the canonical agent definition of Wooldridge and Jennings~\cite{wooldridge1995intelligent}, applying it at the whole-robot granularity. A recent comprehensive survey on LLM-based autonomous agents~\cite{wang2024llmagents} identifies coordination, planning, and tool use as key capabilities---precisely the capabilities FSAR governs at the fleet level rather than fragmenting within individual robots. Foundational coordination mechanisms such as the Contract Net Protocol~\cite{smith1980contract}, the FIPA Agent Communication Language~\cite{fipa2002acl}, and organizational methodologies like Gaia~\cite{wooldridge2000gaia} and MOISE+~\cite{hubner2007moise} have influenced how agent societies are structured. MOISE+ is particularly relevant as it models organizational roles, groups, and norms---concepts that parallel FSAR's trust scopes and authority assignments, though at a finer agent granularity. Platforms such as JADE~\cite{bellifemine2007jade} provide practical implementations of these ideas. FSAR draws on MAS concepts but applies them at a different granularity. In MAS, agents are often fine-grained (one per function or role); in FSAR, the agent is the whole robot. This coarser granularity is motivated by embodied systems requirements: physical identity persistence, real-time safety, and locally coherent recovery. FSAR's trust model relates to trust and reputation work in MAS, but with a focus on embodied contexts where trust scopes are tied to physical capabilities and safety-relevant actions. The comprehensive survey of multi-agent reinforcement learning by Busoniu et al.~\cite{busoniu2008marl} documents the computational complexity that arises when multiple learning agents must coordinate---exponential state-action spaces, non-stationarity, and credit assignment difficulties---providing empirical motivation for FSAR's single-agent design: by keeping each robot as one coherent agent, FSAR avoids the combinatorial coordination overhead that multi-agent RL systems must manage.

\subsection{Middleware, Discovery, and Runtime Governance}
\label{sec:related:middleware}

Middleware frameworks such as ROS~2 and DDS-based layers~\cite{quigley2009ros,macenski2022ros2,pardo2003dds,brugali2009component}, and robot operating systems like RoboOS~\cite{tan2025roboos}, provide multi-robot communication infrastructure. FSAR operates above this level, focusing on governance rather than connectivity. Distributed consensus theory~\cite{olfatisaber2007consensus} provides the mathematical foundations for agreement protocols in networked agents; FSAR's policy composition layer draws on similar ideas but applies them to capability-level governance decisions rather than continuous-state consensus. The SROS2 security framework~\cite{mayoralvilches2022sros2} addresses trust management at the ROS~2 communication level through access control policies and encrypted transport; FSAR's trust model operates at a higher abstraction layer---scoping trust to capability delegation rather than message-level access---but a production implementation could leverage SROS2 as the underlying transport security layer. Work on service discovery, capability sharing~\cite{zlot2006market,dastani2005programming,macenski2023ros2survey}, and capability-based ontologies such as KnowRob~\cite{tenorth2013knowrob} relates to the shared ECM registry, but FSAR treats discovery as governed and delegation as conditional rather than as a neutral directory. Work on runtime governance and safety-oriented control~\cite{fisher2013verifying,luckcuck2019formal}, including the autonomic computing vision~\cite{kephart2003autonomic}, shares FSAR's emphasis on layered recovery, though FSAR extends it from single-robot to fleet coordination. Our prior work~\cite{meyer1992applying,bures2006sofa,qin2025paper5} introduced ECM contracts and release discipline; Paper~6 extends that agenda to fleet-scale coordination where version visibility, contract compatibility, and upgrade awareness become fleet operations concerns.

\subsection{Human-Robot Interaction and Supervision}
\label{sec:related:hri}

Work on supervisory control of robot teams, adjustable autonomy, and shared authority~\cite{goodrich2007human,sheridan2016humanrobot,chen2014supervisory,parasuraman2000model,beer2014toward,endsley1995adjustable} is relevant to FSAR's hierarchical human supervision model. Foundational work on adjustable autonomy~\cite{dorais1999adjustable} established the principle that human-robot authority should be dynamically reconfigurable rather than fixed at design time---a principle FSAR extends to the fleet level through its hierarchical supervision model ($H_i$ for local, $H_F$ for fleet-level oversight) with runtime-adjustable supervision density. FSAR's contribution is the distinction between local and fleet-level supervision with well-defined escalation paths, structuring oversight hierarchically to match fleet coordination structure.

\section{Conclusion}
\label{sec:conclusion}

This paper introduced Federated Single-Agent Robotics (FSAR), a runtime architecture for multi-robot coordination built on the principle that each robot should remain a coherent single embodied agent. Rather than fragmenting robots internally into multi-agent societies to enable fleet coordination, FSAR achieves coordination through federation across intact robot runtimes.

We formalized the model and coordination semantics, described a fleet runtime architecture, and evaluated the design against centralized and decomposition-heavy baselines across five scenarios with scaling experiments up to 16 robots. Results provide evidence that federated single-agent coordination preserves task completion while improving governance locality, recovery containment, and audit attributability.

The central finding is that multi-robot coordination does not require intra-robot multi-agent fragmentation. Fleet coordination can emerge from federation across coherent robot runtimes rather than from decomposition within them. More broadly, future embodied operating systems could benefit from being understood as federated fleet operating environments. The federated fleet setting exposes evaluation dimensions not captured by current benchmarks---including cross-robot authority assignment, fleet-level recovery containment, and hierarchical human oversight---whose development is the subject of our forthcoming work.

\appendix

\section{Self-Contained Foundations from Papers~2--4}
\label{app:foundations}

Papers~2--4 in this series are available as arXiv preprints~\cite{qin2025paper2,qin2025paper3,qin2025paper4}. To ensure Paper~6 is self-contained, we provide the key definitions inherited from those papers that are used in the present work.

\paragraph{Policy scope and composition (Paper~2).} Each robot $r_i$ maintains a local policy scope $P_i$ that governs which capabilities may be activated, under what conditions, and with what constraints. A policy $P_i$ maps capability--context pairs to admission decisions: $P_i(e, \sigma) \in \{\texttt{allow}, \texttt{deny}, \texttt{review}\}$. Paper~6 extends this to cross-robot policy composition: when $r_i$ requests capability $e$ from $r_j$, both $P_i$ (requester policy) and $P_j$ (executor policy) must jointly admit the action, formalized as $\Pi(P_i, P_j, e, \sigma) \neq \texttt{deny}$ (invariant \textbf{I3}).

\paragraph{Recovery levels and degradation (Paper~3).} Recovery in a single-agent runtime follows a four-level hierarchy: (1)~local retry within the executing runtime, (2)~peer-assisted recovery via capability delegation, (3)~fleet-level reassignment, and (4)~human escalation. Each level has a budget $(t_{\max}, n_{\max})$ constraining retry duration and count before escalation. Paper~6 preserves this hierarchy and adds the monotone escalation invariant (\textbf{I4}): recovery must proceed outward through levels without skipping.

\paragraph{Human oversight layers (Paper~4).} Each robot has a local human supervisor $H_i$ responsible for safety overrides, policy exceptions, and anomaly review at the single-robot level. Paper~6 introduces fleet-level supervision $H_F$ and defines the hierarchical oversight structure $\Lambda = (\{H_i\}, H_F)$, with escalation from local to fleet supervision triggered by cross-robot dependencies that exceed local scope.

\section*{Data availability}
The simulator code, baseline implementations, evaluation scripts, and random seeds used in this study are publicly available at \url{https://github.com/s20sc/fsar-fleet-coordination}.

\printcredits

\bibliographystyle{elsarticle-num-names}
\bibliography{references}

\end{document}